\documentclass[preprint,12pt]{elsarticle}

\usepackage{amssymb}
\usepackage{multirow}
\usepackage{colortbl}
\usepackage{xcolor}
\usepackage{subcaption}

\journal{ArXiv}

\begin{document}

\begin{frontmatter}



\title{The use of Multi-domain Electroencephalogram Representations in the building of Models based on Convolutional and Recurrent Neural Networks for Epilepsy Detection}


\author[inst1]{Luiz Antonio Nicolau Anghinoni}

\affiliation[inst1]{Graduate Program in Electrical Engineering, Federal University of Technology – Paraná,
            addressline={Via do Conhecimento, Km 1, Fraron}, 
            city={Pato Branco},
            postcode={85503-390}, 
            state={Paraná},
            country={Brazil}}

\author[inst1]{Gustavo Weber Denardin}

\author[inst2]{Jadson Castro Gertrudes}

\affiliation[inst2]{organization={Computer Science Department, Federal University of Ouro Preto},
            addressline={Rua Professor Paulo Magalhães Gomes, 122, Bauxita}, 
            city={Ouro Preto},
            postcode={35400-000}, 
            state={Minas Gerais},
            country={Brazil}}

\author[inst1]{Dalcimar Casanova}

\author[inst1]{Jefferson Tales Oliva}

\begin{abstract}
Epilepsy, affecting approximately 50 million people globally, is characterized by abnormal brain activity and remains challenging to treat. The diagnosis of epilepsy relies heavily on electroencephalogram (EEG) data, where specialists manually analyze epileptiform patterns across pre-ictal, ictal, post-ictal, and interictal periods. However, the manual analysis of EEG signals is prone to variability between experts, emphasizing the need for automated solutions. Although previous studies have explored preprocessing techniques and machine learning approaches for seizure detection, there is a gap in understanding how the representation of EEG data (time, frequency, or time-frequency domains) impacts the predictive performance of deep learning models. This work addresses this gap by systematically comparing deep neural networks trained on EEG data in these three domains. Through the use of statistical tests, we identify the optimal data representation and model architecture for epileptic seizure detection. The results demonstrate that frequency-domain data achieves detection metrics exceeding 97\%, providing a robust foundation for more accurate and reliable seizure detection systems.
\end{abstract}




\end{frontmatter}


\section{Introduction}\label{section:INTRODUCTION}

Neurophysiological monitoring has become one of the most important field of research due to its ability to detect and analyze the brain's electrical activity \cite{chaddad2023electroencephalography}. This important role has led researchers to develop various methods for gathering information about brain activity, resulting in significant advancements in medical signal and image acquisition systems \cite{freeman2012imaging}. Among these advancements are functional neuroimaging techniques, such as functional magnetic resonance imaging, magnetoencephalography (MEG), positron emission tomography (PET), and electroencephalography \cite{freeman2012imaging}.

Among these techniques, electroencephalography stands out due to three key advantages: it is a non-invasive method that allows data generation from any individual, has excellent temporal resolution—effectively capturing events occurring within milliseconds—and is relatively cost-effective compared to other examinations \cite{algumaei2023neuroscience}.

Electroencephalography monitors the brain's electrical activity through electrodes placed on the scalp, and the resulting data, known as the electroencephalogram (EEG), consists of a time series of electrical potentials that reflect neurological activity \cite{muller2020electroencephalography}. The EEG signal is widely used in the field of neuroscience and has the potential to advance brain–computer interfaces \cite{aggarwal2022review}, facilitate emotion detection \cite{gannouni2021emotion}, enable classification of sleep stages \cite{eldele2021attention} and help clinicians and researchers in identifying brain diseases, including but not limited to Alzheimer’s disease \cite{safi2021alzearly}, dyslexia \cite{ortiz2020dyslexia}, schizophrenia \cite{siuly2020schizophrenia}, Creutzfeldt–Jakob disease \cite{appel2021creutzfeldt} and cognitive impairment \cite{alvi2022mild}.

Epilepsy, for example, is a neurological disorder characterized by abnormal brain activity that can lead to seizures, unusual behaviors, or even loss of consciousness. This illness remains one of the most common and challenging brain disorders to treat, affecting around 50 million people globally, with approximately 40\% not responding to anti-seizure medications \cite{epilepsy2019public}. The diagnosis of epilepsy relies on clinical information, patient history, observation of seizures, and neuroimaging techniques, such as EEG \cite{shin2014review}.

In practice, experts visually examine EEG data to identify epileptiform patterns, which are classified into four distinct periods that aid in assessing the disorder’s progression and characteristics \cite{fisher2014can}. The epileptiform patterns are:

\begin{itemize}
\item \textbf{Pre-ictal:} neurological activity prior to seizure.
\item \textbf{Ictal:} period when the seizure occurs.
\item \textbf{Post-ictal:} neurological activity shortly after the seizure.
\item \textbf{Interictal:} abnormal neurological activity between seizures
\end{itemize}

Despite the great usefulness of EEG in the epilepsy diagnosis, the large volume of data makes its manual analysis by healthcare professionals difficult and time-consuming \cite{ de2023feature}. Furthermore, the interpretation of EEG signals can vary between specialists, even when analyzing the same patterns \cite{oliva2019classification}.

While the literature has explored various preprocessing techniques and machine learning methods, an important gap remains in understanding how the representation of EEG data (whether in the time, frequency, or time-frequency domains) affects the performance of predictive models in epilepsy diagnosis.

In this work, we hypothesize that the domain of EEG representation significantly influences the predictive performance of deep learning models in detecting epileptic seizures. By systematically evaluating and comparing these representations, this study aims to identify the most effective domain for seizure detection. The objective is to enhance seizure detection and provide better medical support in diagnosing epilepsy and assisting decision-making processes.

\section{Electroencephalography}
\label{section:EEG}

\begin{figure}
    \centering
    \includegraphics[width=0.5\linewidth]{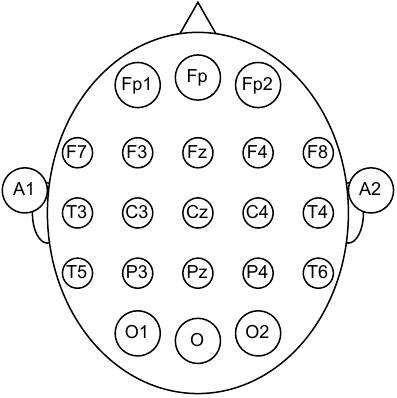}
    \caption{Example of electrode arrangement on the scalp and their nomenclature \cite{freeman2012imaging}.}
    \label{fig:local_eletrodos}
\end{figure}

The EEG is a type of exam used to measure the potential difference generated by the activity of neurons in the brain. The exam involves measuring this potential using electrodes positioned symmetrically on the scalp considering the international 10--20 system method \cite{ein2023eeg}. The scalp places are named according to the lobe region, such as central (C), frontal (F), frontal polar (Fp), occipital (O), parietal (P), and temporal (T). Odd numbers in electrode labeling correspond to the left brain hemisphere, even values are related to the right brain hemisphere, and $z$ denotes the midline region. The reference electrodes (A1 and A2) are placed at the ears \cite{ein2023eeg}. Figure \ref{fig:local_eletrodos} shows an example of the arrangement of the electrodes and their nomenclature.

The EEG, which consists of multiple signals (also known as channels), is formed by the potential difference between pairs of nearby electrodes. Each channel in the EEG represents the activity in a specific brain region based on the potential measured between two electrodes over time \cite{shoeb2009application}. For example, the channel $Fp1-F7$ is formed by the potential difference between electrodes $Fp1$ and $F7$, representing neural activity in the frontal lobe of the left hemisphere \cite{shoeb2009application}.

\begin{figure}[htbp]
    \centering
    \begin{subfigure}{0.49\textwidth}
        \centering
        \includegraphics[width=\linewidth]{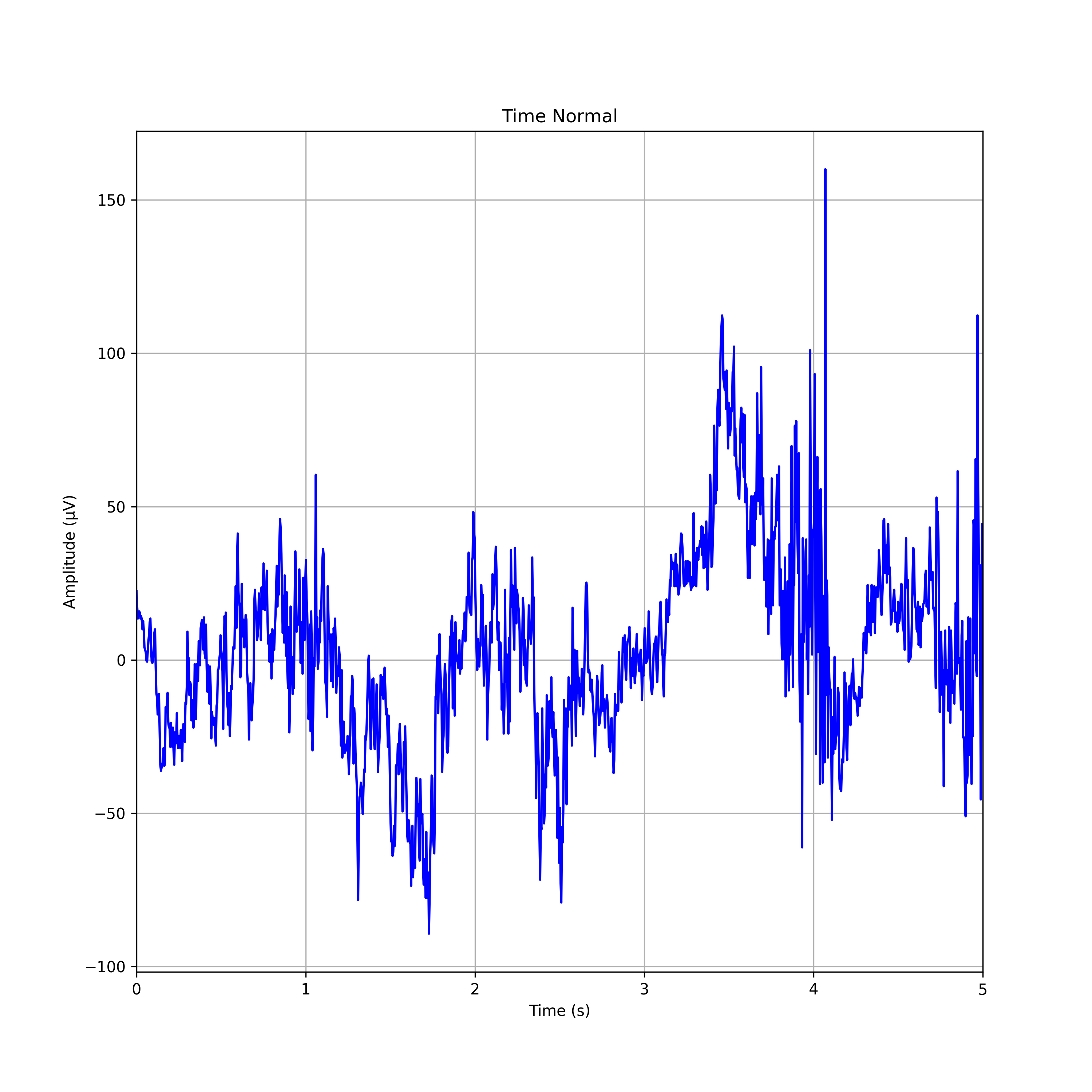}
        \caption{Normal time signal}
        \label{fig:time_normal}
    \end{subfigure}
    \hfill
    \begin{subfigure}{0.49\textwidth}
        \centering
        \includegraphics[width=\linewidth]{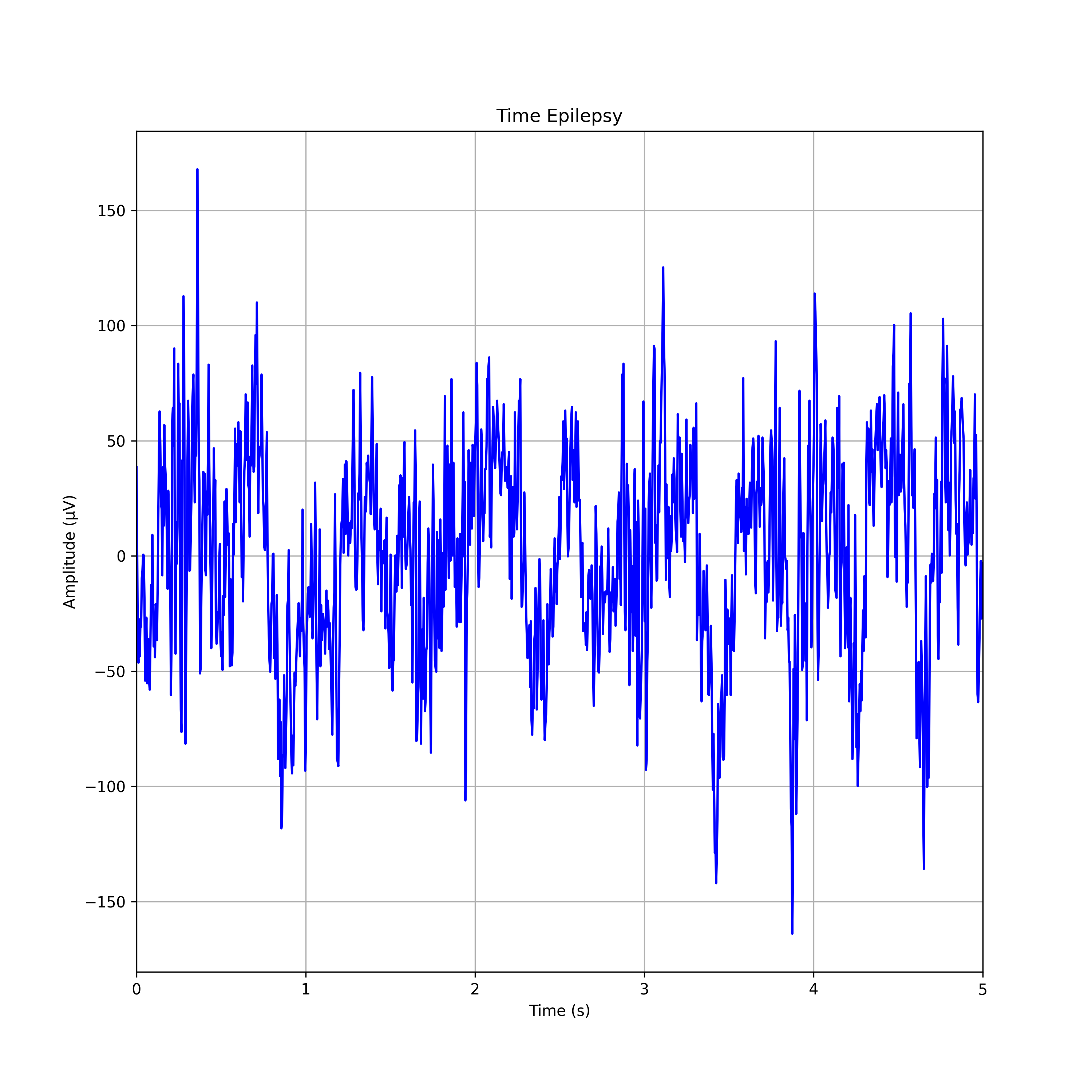}
        \caption{Epileptic time signal}
        \label{fig:time_epileptic}
    \end{subfigure}
    \caption{Time signal}
    \label{fig:time-signal}
\end{figure}

Figure \ref{fig:time-signal} shows an example of data in time domain, where the left graph represents a normal signal, and the right graph represents an epileptic seizure signal.

\section{EEG Spectral Features}
\label{section:spectral}

As amplitude oscillations represent EEG in time domain, it is possible to decompose this signal into frequency domain \cite{jansen1988structuraleeg}. This decomposition can be achieved through specific methods such as the Fourier Transform (FT) for continuous signals or the Discrete Fourier Transform (DFT) \cite{sundararajan2001discrete} for finite discrete signals. Given that EEG recordings are both discrete and finite, the DFT is particularly well-suited for analyzing these signals. Mathematically, the DFT is represented by Equation \ref{eq:dft}, where $X[k]$ denotes the $k$-th Fourier component, $j$ is the imaginary unit, and $2\pi k$ represents the angular frequency. The DFT can be computed using the Fast Fourier Transform (FFT) algorithm \cite{brigham1988fastfourier}, which significantly enhances computational efficiency of the DFT \cite{akin2002comparison}.

\begin{equation}
X[k] = \sum_{n=0}^{N-1} x[n] \cdot  e^{\frac{-j2\pi kn}{N}}
\label{eq:dft}
\end{equation}

The application of the FFT in an EEG allowed the identification of fundamental features of the signal that are often not evident in temporal representations \cite{freeman2012imaging}. Through this transformative process, it was possible to observe the correlation between some frequency bands with different brain states, functions, and pathologies. These frequency bands, also known as base waves, are patterns of electrical activity generated by the brain in different states of consciousness \cite{khosla2020comparative}, and have been classified as \cite{newson2019bandwaves}:

\begin{itemize}
    \item \textbf{Delta} ($0.5$ – $3.5 Hz$): They are characteristic of deep sleep stages.
    \item \textbf{Theta} ($3.5$ – $7.5 Hz$): They are typical during deep sleep. They play an important role in infancy and childhood. In the awake adult, high theta activity is considered abnormal and it is related to brain disorders, such as epilepsy.
    \item \textbf{Alpha} ($7.5$ – $12.5 Hz$): They appear spontaneously in normal adults during wakefulness, under relaxation and mental inactivity conditions.
    \item \textbf{Beta} ($12.5$ – $30 Hz$): They are enhanced upon mental calculations, expectancy or tension over the entire surface of the scalp.
    \item \textbf{Low Gamma} ($30$ – $60 Hz$): They are associated with different sensory and cognitive processes.
    \item \textbf{High Gamma} ($80$+$Hz$): They are associated with chattering action potentials.
\end{itemize}

The FFT results allow for computing the energy distribution across frequency bands, yielding the Power Spectral Density (PSD). This representation shows the energy as a function of frequency and allows the quantification of the intensity of each band in the signal \cite{freeman2012imaging}. The PSD can be generated by Equation \ref{eq:psd}, where $X[k]$ are the frequency components from the FFT. 

\begin{equation}
PS[k] = \left|X[k]\right|^2
\label{eq:psd}
\end{equation}

\begin{figure}[htbp]
    \centering
    \begin{subfigure}{0.49\textwidth}
        \centering
        \includegraphics[width=\linewidth]{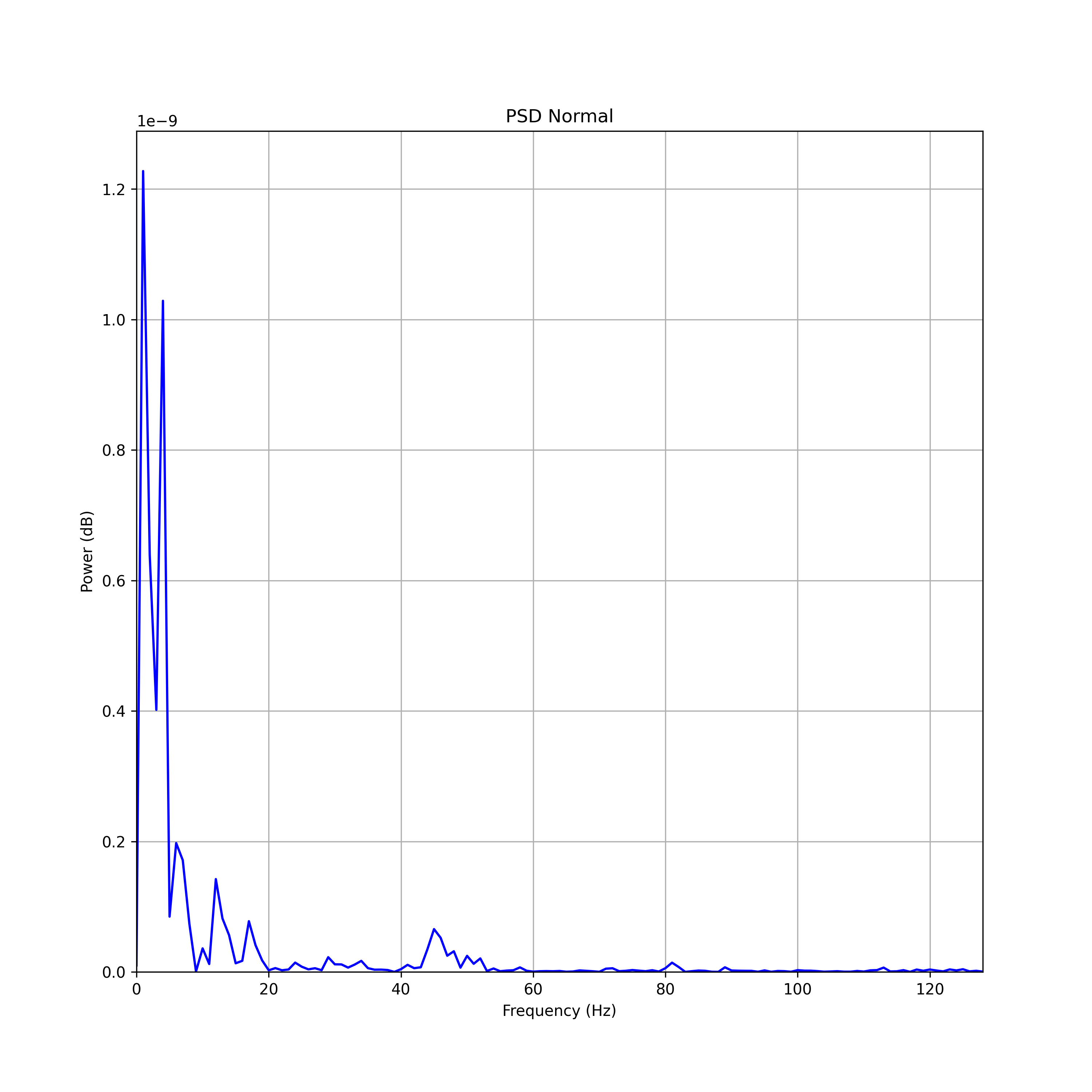}
        \caption{Normal PSD Welch signal}
        \label{fig:welch_normal}
    \end{subfigure}
    \hfill
    \begin{subfigure}{0.49\textwidth}
        \centering
        \includegraphics[width=\linewidth]{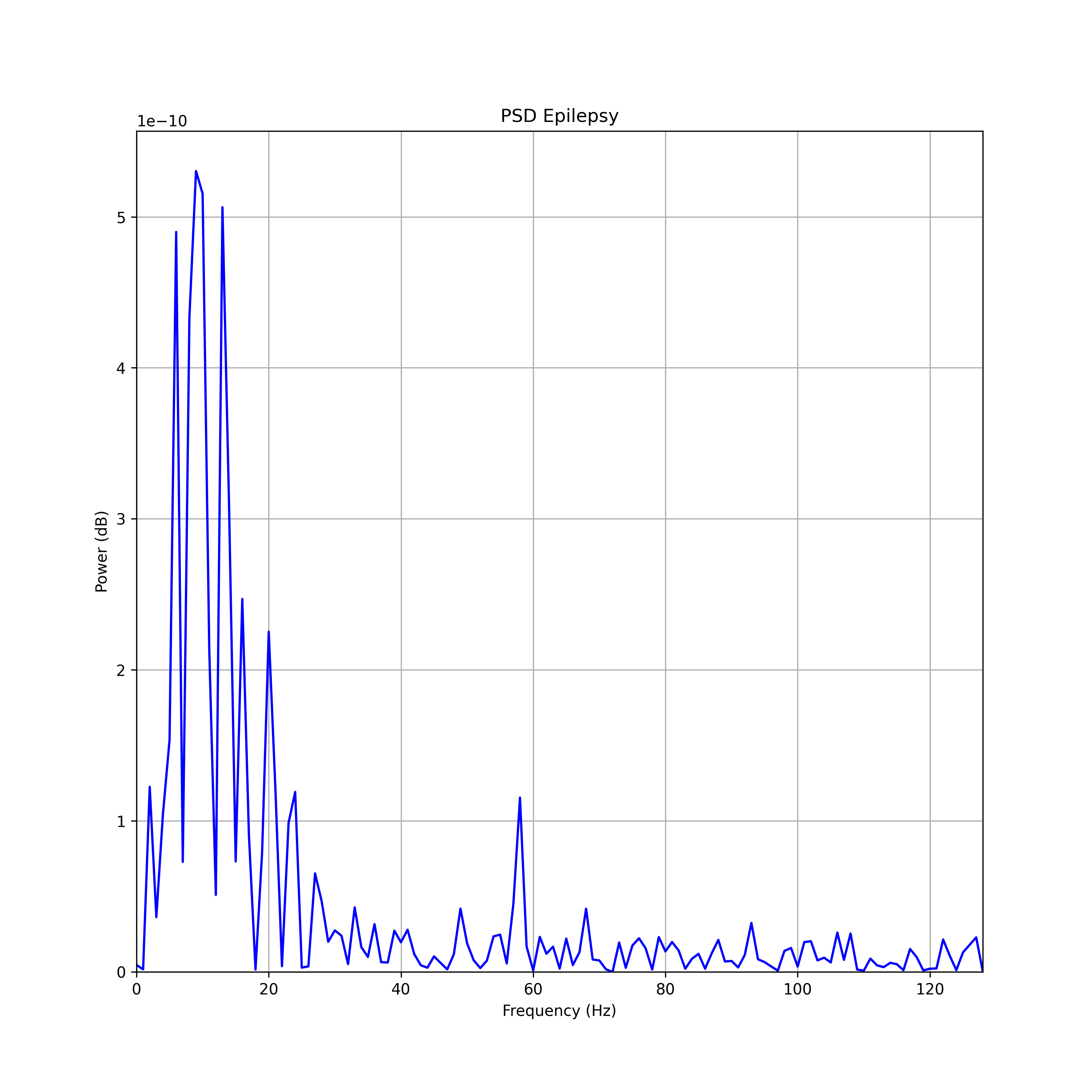}
        \caption{Epileptic PSD Welch signal}
        \label{fig:welch_epileptic}
    \end{subfigure}
    \caption{PSD Welch signal}
    \label{fig:psdwelch-signal}
\end{figure}

The FFT is used to get the spectral estimate over the entire signal but it is sensitive to non stationary signals \cite{zhang2023applied}. If we want a better estimate for signal with non stationary components, it is recommend to use the Welch method \cite{welch1967use}. This method splits the original signal into $N$ overlapping consecutive segments using Hamming window's \cite{podder2014comparative} and then calculates the PSD of each segment. Welch’s method is one of the most widely used power spectrum analysis methods because it reduces the influence of boundary effects on the PSD, providing a more stable results than FFT \cite{zhang2023applied}. Figure \ref{fig:psdwelch-signal} illustrates the PSD computed using the Welch method, where the left graph represents a normal signal, and the right graph represents an epileptic seizure signal.

Although broadly used, separate analyses in time and frequency domains may not provide complete signal information, limiting an automatic analysis for diagnostic purposes \cite{martinez2012timefreq}. In this context, the time-frequency domain can be considered for signal representation. In this domain, the signals can be simultaneously analyzed in time and frequency.

For the time-frequency domain representation, the short-time Fourier transform (STFT) can be applied. The STFT consists in chopping the data into pieces and then calculating the PSD for each piece or using a time-evolving window to focus at different segments of data \cite{gabor1946theory}. The Equation \ref{eq:stft} represents the STFT, where $X[n,k]$ are the function in time and frequency, $n$ is the location of the window along time axis, $k$ is the frequency bin, $N$ represents the window length and $h$ is the window function.

\begin{equation}
X[n,k]=\sum_{n=0}^{N-1}x[n]h[n-\tau]e^{\frac{-j2\pi kn}{N}}
\label{eq:stft}
\end{equation}

\subsection{Multitaper in Spectral Features}

Signal analysis in spectral domains is commonly performed by variations of the Fourier transform, such as DFT (frequency domain) and STFT (time-frequency domain). These conventional Fourier transforms has some limitations, as it can generate biased spectral components, making it unreliable for certain analyses \cite{oliva2021binary}.

To address this issue, the Multitaper \cite{thomson1982spectrum} method was proposed. It is a modified Welch’s method that provides features similar to those of the STFT and Welch’s methods, but its stability is improved and the number of parameters to be determined is reduced as it uses multiple tapers for superposition \cite{zhang2023applied}.

The Multitaper method can be computed in a time domain signal $S$ with length $N$ using Equation \ref{eq:multitaper}, where $Y_{k,l}$ is the $k$-th spectral component generated by using the $l$-th data taper ($M$), $2\pi k$ is the angular frequency, $j$ is the imaginary unit, and $e^{\frac{-j2\pi kn}{N}}$ is equivalent to the Euler’s complex exponential function \cite{moskowitz2002course}.

\begin{equation}
Y_{k,l} = \sum_{n=1}^{N-1}M_{n,l} S_ne^{\frac{-j2\pi kn}{N}}
\label{eq:multitaper}
\end{equation}

This equation is a cross-correlation operation between a signal represented in time-domain and the Euler’s function, including $k$-tapers generated through Slepian sequences, which originate from spectral concentration problem  \cite{oliva2021binary, slepian1978prolate}. To generate a one-dimensional set of frequency components $X$ with reduced broadband and narrow-band bias \cite{babadi2014review}, the matrix $Y$ is averaged as shown in Equation \ref{eq:multitaper-matrix}, where $q$ is the number of data tapers used for signal processing.

\begin{equation}
X_k=\frac{1}{q}\sum_{l=0}^{q-1}Y_{k,l}
\label{eq:multitaper-matrix}
\end{equation}

\begin{figure}[htbp]
    \centering
    \begin{subfigure}{0.49\textwidth}
        \centering
        \includegraphics[width=\linewidth]{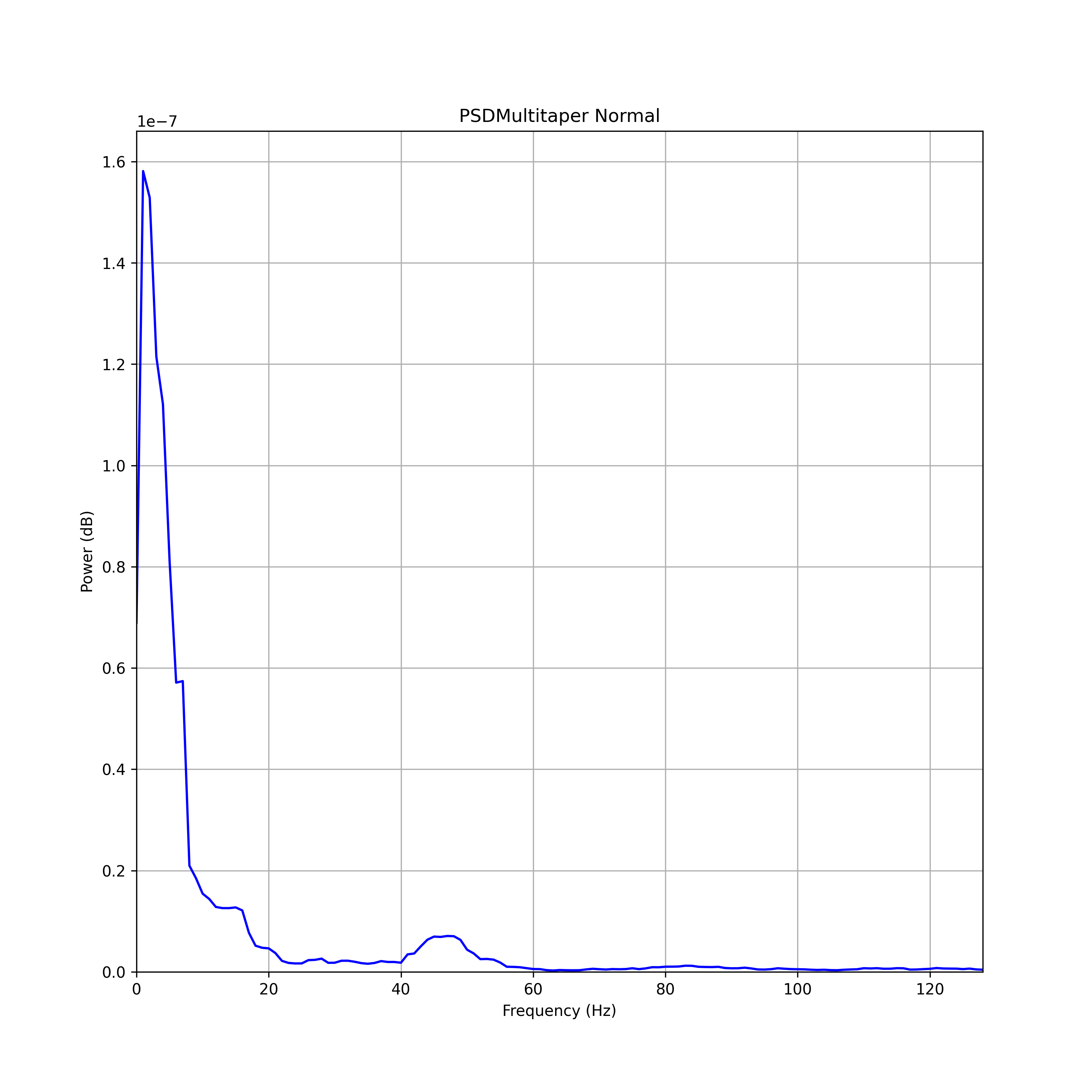}
        \caption{Normal PSD Multitaper signal}
        \label{fig:multitaper_normal}
    \end{subfigure}
    \hfill
    \begin{subfigure}{0.49\textwidth}
        \centering
        \includegraphics[width=\linewidth]{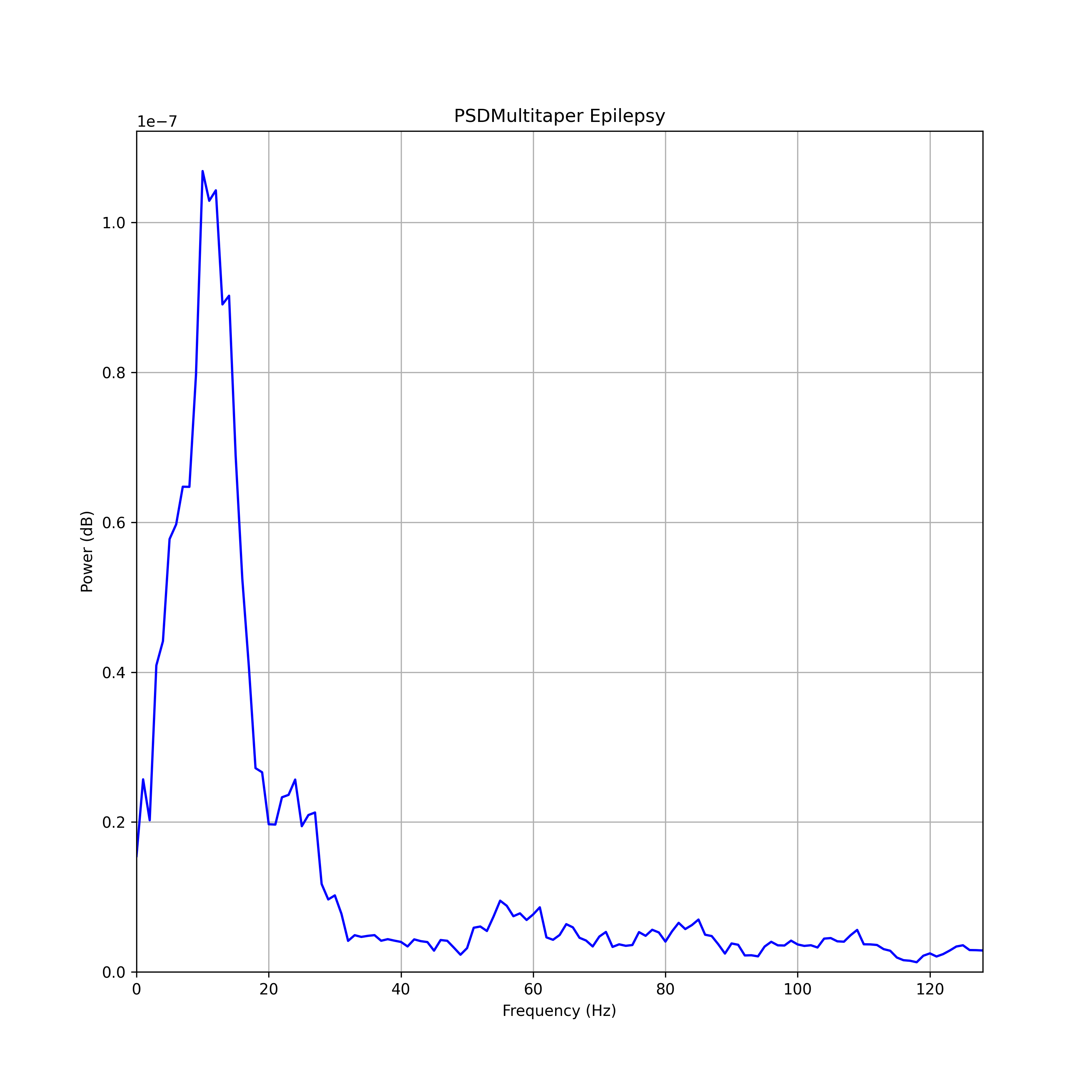}
        \caption{Epileptic PSD Multitaper signal}
        \label{fig:multitaper_epileptic}
    \end{subfigure}
    \caption{PSD Multitaper signal}
    \label{fig:psdmultitaper-signal}
\end{figure}

The Multitaper method can transform a time-domain signal into representations in different domains. For PSD data, it can be computed by applying Equation \ref{eq:psd} in the set of frequency components $X$ generated by Equation \ref{eq:multitaper-matrix}. Figure \ref{fig:psdmultitaper-signal} illustrates a PSD data computed using the Multitaper method, where the left graph represents a normal signal, and the right graph represents an epileptic seizure signal.

\begin{figure}[htbp]
    \centering
    \begin{subfigure}{0.49\textwidth}
        \centering
        \includegraphics[width=\linewidth]{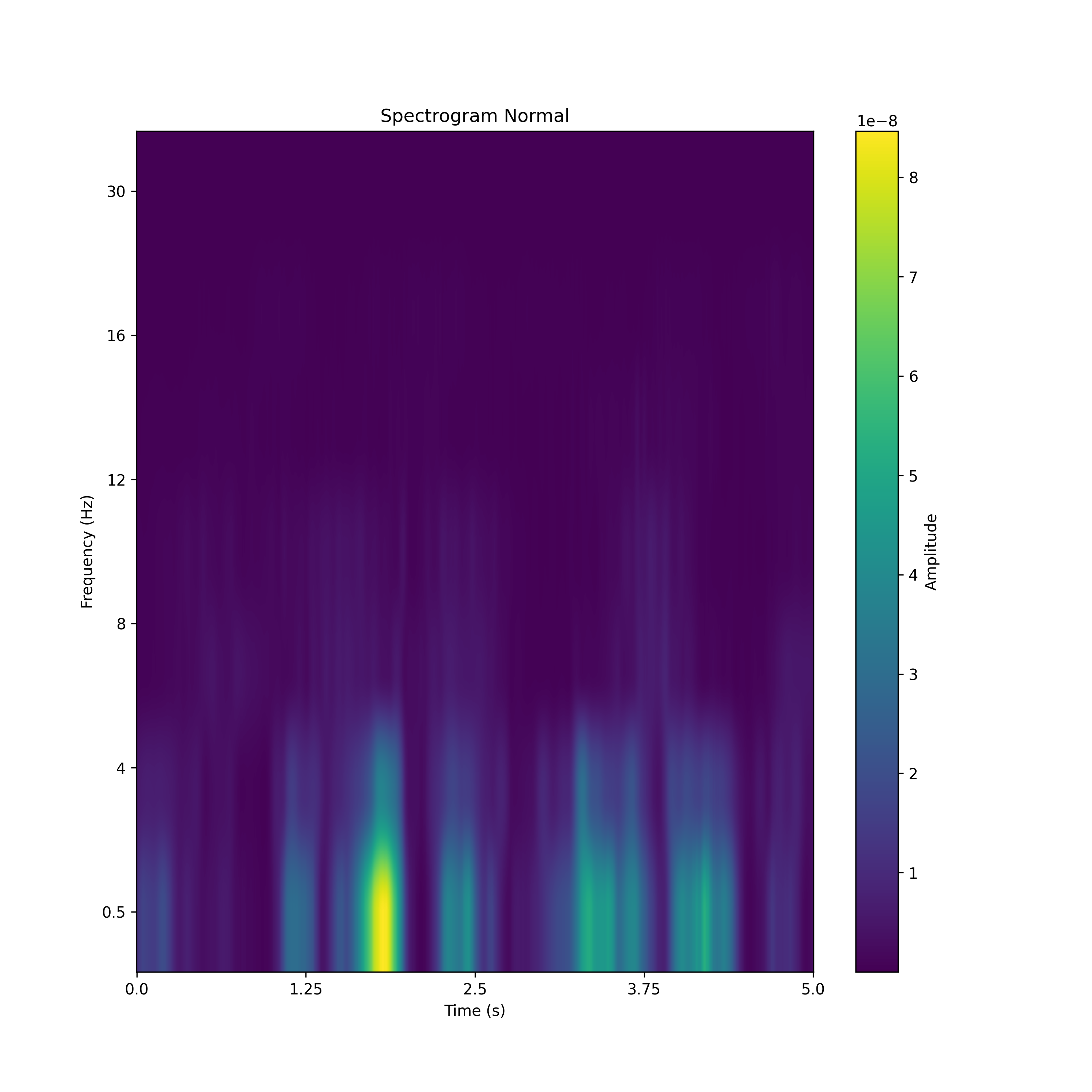}
        \caption{Normal spectrogram signal}
        \label{fig:spec_normal}
    \end{subfigure}
    \hfill
    \begin{subfigure}{0.49\textwidth}
        \centering
        \includegraphics[width=\linewidth]{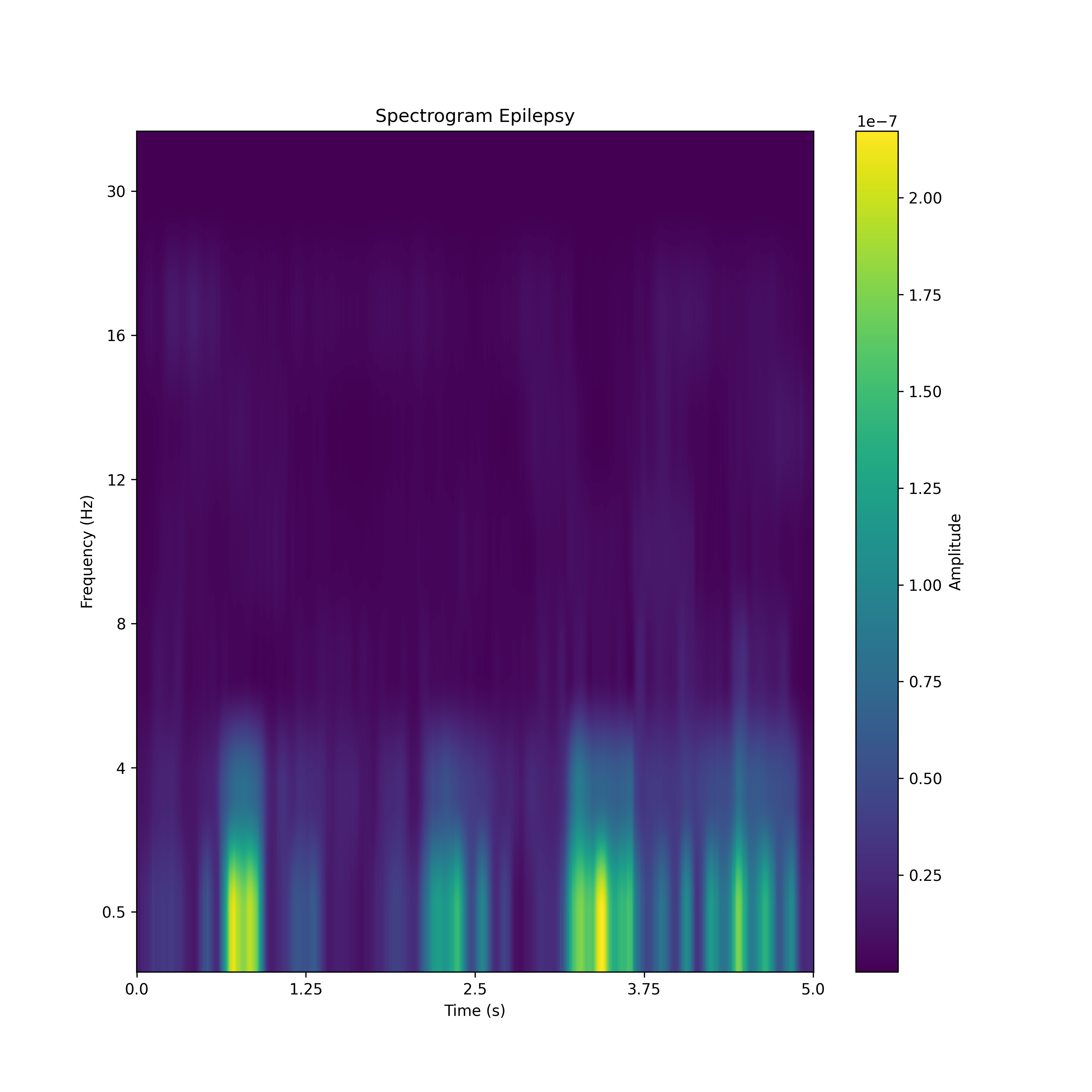}
        \caption{Epileptic spectrogram signal}
        \label{fig:spec_epileptic}
    \end{subfigure}
    \caption{Spectrogram signal}
    \label{fig:spec-signal}
\end{figure}

For the time-frequency domain, a spectrogram can be generated by adapting Equation \ref{eq:multitaper} and Equation \ref{eq:multitaper-matrix}. In this adaptation, defined by Equation \ref{eq:multitaper-sg}, time ($t$), window length ($w$) and  the number of tapers of data ($q$) used for processing signal must be considered. Thus, in comparison with STFT, which uses a window function to traverse signals, the adapted Multitaper employs the data tapers, with length $w$, as sliding windows \cite{oliva2018differentiation}. Figure \ref{fig:spec-signal} represents the data in time-frequency domain using the Multitaper method, where the left graph represents a normal spectrogram, and the right graph represents an epileptic seizure spectrogram.

\begin{equation}
{SG}_{t,k}=\left|\frac{1}{q}\sum_{l=0}^{q-1}\sum_{n=t}^{t+w-1}M_{n-t+1,l}S_n{\frac{-j2\pi kn}{N}}\right|^2
\label{eq:multitaper-sg}
\end{equation}

\section{Related Works}
\label{section:related}

With the arrival of machine learning, several works over the years have presented proposals to automate the detection or prediction of epileptic seizures using EEG data. These existing approaches can be divided into two categories \cite{alotaiby2014eeg}. The first category refers to seizure detection, which aims to assist healthcare professionals in diagnosing illnesses. The second category focuses on seizure prediction to enable early intervention to reduce the risks associated with seizures and improve the patient's quality of life.

The primary distinctions among the analyzed studies lie in the domain representation of EEG data and the training approach of the models. EEG data for machine learning applications is typically represented in the time, frequency, and/or time-frequency domains \cite{craik2019deep}. The choice of domain significantly affects the model's ability to capture relevant features, as well as the computational complexity, with frequency or time-frequency representations often requiring more processing power but yielding more accurate predictions.

Commonly used models to process these representations include convolutional neural networks (CNNs) \cite{li2021survey}, recurrent neural networks (RNNs) \cite{yu2019review}, and hybrid architectures combining these approaches (CRNNs) \cite{xu2019recurrent}.

Another critical distinction among these works is their approach to training, specifically whether they employ patient-specific or generic models. Patient-specific models \cite{zhou2018epileptic, zhang2022epileptic, jana2021deep, liu2023epileptic} are tailored to individual patients by using their own EEG data for training. These models often achieve higher detection accuracy by capturing unique characteristics of a patient’s brain activity. However, such models increases the risk of overfitting and reduces their generalizability to unseen data.

In contrast, generic models \cite{thodoroff2016learning, sharan2020epileptic, kaziha2020convolutional, thuwajit2021eegwavenet, alharthi2022epileptic, singh2022two, hu2023exploring} are trained on datasets encompassing multiple patients, enabling them to generalize better across diverse populations. This makes generic models more suitable for real-world applications.

Consistent preprocessing steps, such as channel selection (manual or algorithmic), segmentation into smaller windows, the use of overlapping windows, and noise/artifact filtering, are commonly employed across studies to enhance data quality and model performance. However, many studies lack a clear justification for these choices, particularly regarding window sizes, overlap proportions and filter selection.

For example, \cite{sharan2020epileptic} applies a filter to remove frequencies above 45 Hz, while \cite{thuwajit2021eegwavenet} uses a low-pass filter with a cutoff at 64 Hz. Similarly, \cite{liu2023epileptic} employs a bandpass filter ranging from 0.1 to 40 Hz, and \cite{hu2023exploring} utilizes a Butterworth bandpass filter between 5 and 50 Hz. These examples illustrate the diversity of filtering techniques used in seizure detection studies, often without a clear justification for their selection or an analysis of their impact on EEG data.

For this review, we focused on studies using the same dataset as this work to ensure a fair comparison and includes only methods employing deep learning models.

Early studies have focused on transforming EEG signals into the frequency domain or time-frequency-domain. In \cite{thodoroff2016learning}, spectral images were generated from EEG frequency bands and used to train a CRNN model, achieving a sensitivity of 85\%. Similarly, \cite{zhou2018epileptic} compared the effectiveness of time-domain and frequency-domain representations, demonstrating that a CNN trained on FFT-transformed data significantly outperformed one trained on raw EEG signals (93\% vs. 47.9\% accuracy, respectively). Expanding on these approaches, \cite{sharan2020epileptic} compared EEG representations using FFT and wavelet transform (WT) \cite{daubechies1990wavelet}, segmenting signals into 2-second windows. Two CNN models were trained separately for each transformation, with the WT-based model achieving the highest accuracy (97.25\%).

Further advancing these preprocessing techniques, \cite{alharthi2022epileptic} utilized discrete wavelet transform (DWT) to decompose EEG signals for epileptic seizure detection. This approach employed 18 channels, a 10-second sliding window, and a 1-second overlap, with a hybrid model combining 1D-CNN and BiLSTM layers (CRNN). The method achieved a sensitivity of 96.85\%. Similarly, \cite{zhang2022epileptic} presented another DWT-based approach, where EEG recordings were segmented into 4-second non-overlapping windows and transformed using Daubechies-4 wavelets. The resulting five frequency bands were fed into a Bi-GRU network (RNN), which achieved an average sensitivity of 93.89\% and specificity of 98.49\%.

While many studies emphasize feature extraction and data transformation, some approaches explore raw EEG data directly. For example, \cite{kaziha2020convolutional} applied a CNN model to 100-second EEG segments, achieving 96.74\% accuracy, with a specificity of 100\% and a sensitivity of 82.35\%.
Similary, \cite{thuwajit2021eegwavenet}, a method for automatic seizure detection using a CNN model is proposed, where EEG signals are preprocessed by selecting 21 common channels, and segmenting the data into 4-second windows with a 1-second overlap, achieving a F1-Score of 96.94\%. In \cite{yao2021robust}, they employed a BiLSTM model trained on 23-second non-overlapping EEG segments, achieving an average accuracy of 87.93\%. These results highlight the ability of deep learning models to extract relevant patterns directly from raw signals without explicit transformations.

Beyond differences in data preprocessing, another critical factor in model performance is the choice of window size. \cite{jana2021deep} examined this effect by training CNN models on EEG segments of varying durations (1, 2, 4, and 8 seconds). Their findings suggest that longer windows improve classification accuracy, with 98.81\% and 99.63\% achieved for 4 and 8-second segments, respectively. Similarly, \cite{singh2022two} investigated the impact of window length on seizure detection using LSTM-based models. EEG data from 23 channels were segmented into windows ranging from 5 to 50 seconds, followed by spectral feature extraction using FFT. Their results shows the benefits of longer temporal contexts, with the single-layer LSTM achieving 97.34\% accuracy on 50-second windows and the two-layer LSTM reaching 98.14\% accuracy with 30-second windows.

Besides the works discussed in this section, it is essential to mention that recent research continues to employ classical methods for epileptic seizure detection and prediction, such as the one proposed in \cite{liu2023epileptic}, where the study focuses on using a parametric PSD method to detect and predict epileptic seizures using EEG signals. The raw EEG signals are segmented into 10-second windows with 9-seconds overlap. The Welch method is then applied to compute the PSD, extracting features such as center frequency, power, and bandwidth. These features are classified using traditional algorithms, including Support Vector Machines (SVM), k-Nearest Neighbors (KNN), Decision Trees (DT), and Linear Discriminant Analysis (LDA).

In contrast, modern approaches using transformers have shown promising results for seizure detection, as demonstrated in \cite{hu2023exploring}, where a method for detecting epileptic seizures using a hybrid Transformer model is proposed. Three different types of input representations are evaluated: raw EEG signals, STFT spectrograms, and mixed rhythm signals decomposed using discrete wavelet transform (DWT). The EEG signals are segmented into 5-second windows. The proposed model achieved 91.7\% sensitivity and 0.00\% false positive rate (FPR).

Table \ref{tab::related} summarizes key characteristics of the reviewed methods and contrasts them with this work, considering the following criteria:

\begin{itemize}
    \item \textbf{Feature domain:} indicates whether features were extracted from time domain (TD), frequency domain (FD), time–frequency domain (TFD).
    \item \textbf{Architecture:} refers to the type of neural network model architecture employed in each study.
    \item \textbf{Cross-validation:} technique used to split the dataset for training and testing.
    \item \textbf{Metrics:} measures used to evaluate predictive models.
    \item \textbf{Segment size:} duration (in seconds) of each EEG data segment processed in the study.
    \item \textbf{Statistical test:} indicates whether statistical tests were performed to analyze the results.
\end{itemize}

\begin{table}
\centering
\caption{Comparison among methods published in related literature}
\resizebox{\textwidth}{!}{%
\begin{tabular}{ccccccccc}
Reference & Feature Domain & Architecture & Cross-validation & Metrics & Segment Size & Statistical test \\ \hline
\cite{thodoroff2016learning} & FD & CRNN & LOOCV & SEN & 1 & No \\
\cite{zhou2018epileptic} & TD, FD & CNN & 6-Fold & SEN, SPE, ACC & 1 & No \\
\cite{sharan2020epileptic} & FD, TFD & CNN & 10-Fold & SEN, SPE, ACC, ROC & 2 & No \\
\cite{kaziha2020convolutional} & TD & CNN & 5-Fold & SEN, SPE, ACC & 100 & No \\
\cite{yao2021robust} & TD & RNN & 10-Fold & SEN, SPE, PRE, ACC, ROC & 23 & No \\
\cite{jana2021deep} & TD & CNN & 5-Fold & SEN, SPE, ACC, ROC & 1, 2, 4, 8 & No \\
\cite{thuwajit2021eegwavenet} & TD & CNN & 5-Fold & SEN, SPE, ACC, F1 & 4 & Yes \\
\cite{alharthi2022epileptic} & TFD & CRNN & None & SEN, PRE, ACC & 10 & No \\
\cite{singh2022two} & FD & RNN & 10-Fold & SEN, SPE, ACC, F1 & 5 - 50 (5 in 5)  & No \\
\cite{zhang2022epileptic} & TFD & RNN & 10-Fold & SEN, SPE, ACC & 4 & No \\
\cite{liu2023epileptic} & FD & SVM, KNN, DT, LDA & 10-Fold & SEN, SPE, ACC, MCC & 10 & No \\
\cite{hu2023exploring} & TD, TFD & CNN, Transformer & None & SEN, ACC, F1, FPR & 5 & No \\
\rowcolor{lightgray} This Work & TD, FD, TFD & RNN, CNN, CRNN & 10-Fold & SEN, SPE, PRE, ACC, F1, ROC & 1, 2, 5, 10 & Yes \\ \hline
\end{tabular}%
\label{tab::related}
}
\end{table}

This study addresses gaps in the existing literature by systematically comparing deep neural networks trained on EEG data across three domains: time, frequency, and time-frequency. Through the use of statistical tests and the evaluation of various segment sizes, we provide robust evidence on optimal data representations and architectural choices for epileptic seizure detection.

\section{Methodology}\label{section:MET}

\subsection{Dataset}

The dataset used in this work is the CHB-MIT Scalp EEG Database \cite{guttag2010chb} from PhysioNet. The examinations in this dataset were generated at Boston Children's Hospital. For generation of this database, patients were monitored at a sampling rate of $256 Hz$ with 16-bit quantization, and electrodes were placed following the international 10–20 system for electrode placement. The dataset contains information from 22 patients aged between 1.5 and 22 years old. For each patient, there are between 9 and 42 EEG samples, with durations ranging from 1 to 4 hours. Each sample consists of 23 or more channels. The database records were separated into seizure and non-seizure records and contained a total of 664 EEG files, from which 127 contained one or more occurrences of seizures.  For this study, we used 112 EEG files, as some were found to be corrupted, accounting for a total of 8930 seconds of seizure activity.

The information in each sample is organized into summaries provided along with the dataset, separated by patient. These summaries contain important data regarding the EEG signals, such as record name, file name, start time, end time, number of seizures, seizure start times, seizure end times, number of channels, and channel names. Given the complexity and volume of files, we opted for a database solution to streamline access and enhance the readability of this information.

\subsection{Preprocessing}

\begin{figure}
    \centering
    \caption{Preprocessing pipeline.}
    \includegraphics[height=\textheight]{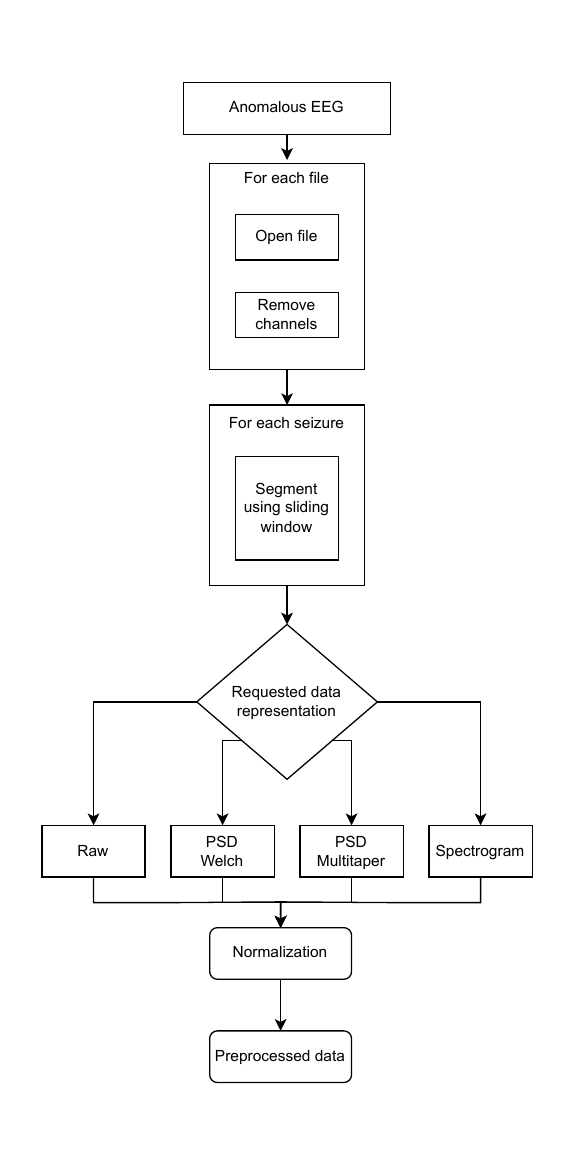}
    \label{fig:preprocess}
\end{figure}

The EEG recordings used from this dataset vary in size, are imbalanced, and exhibit different durations of epileptic seizures. Therefore, preprocessing is necessary to ensure uniform sample length (duration). Figure \ref{fig:preprocess} illustrates the full pipeline of the preprocessing step.

Initially, EEG signals presenting ictal activities (epileptic seizures) are selected from the dataset to reduce data imbalance. This selection is based on information provided in the dataset summary files, which include annotations that map the files containing ictal activities along with their respective start and end times. The selected EEG records are then read, retaining only the channels with common occurrences across all examinations to standardize data input. The retained channels are: FP1-F7, FP1-F3, FP2-F8, FP2-F4, FZ-CZ, F3-C3, F4-C4, F7-T7, F8-T8, C3-P3, C4-P4, CZ-PZ, T8-P8, T7-P7, P7-O1, P3-O1, P8-O2, and P4-O2. In total, 18 channels are selected, while the remaining channels are discarded.

\subsubsection{Sliding Windows}
\label{sub:sliding_window}

\begin{figure}
	\centering
	\includegraphics[width=\textwidth]{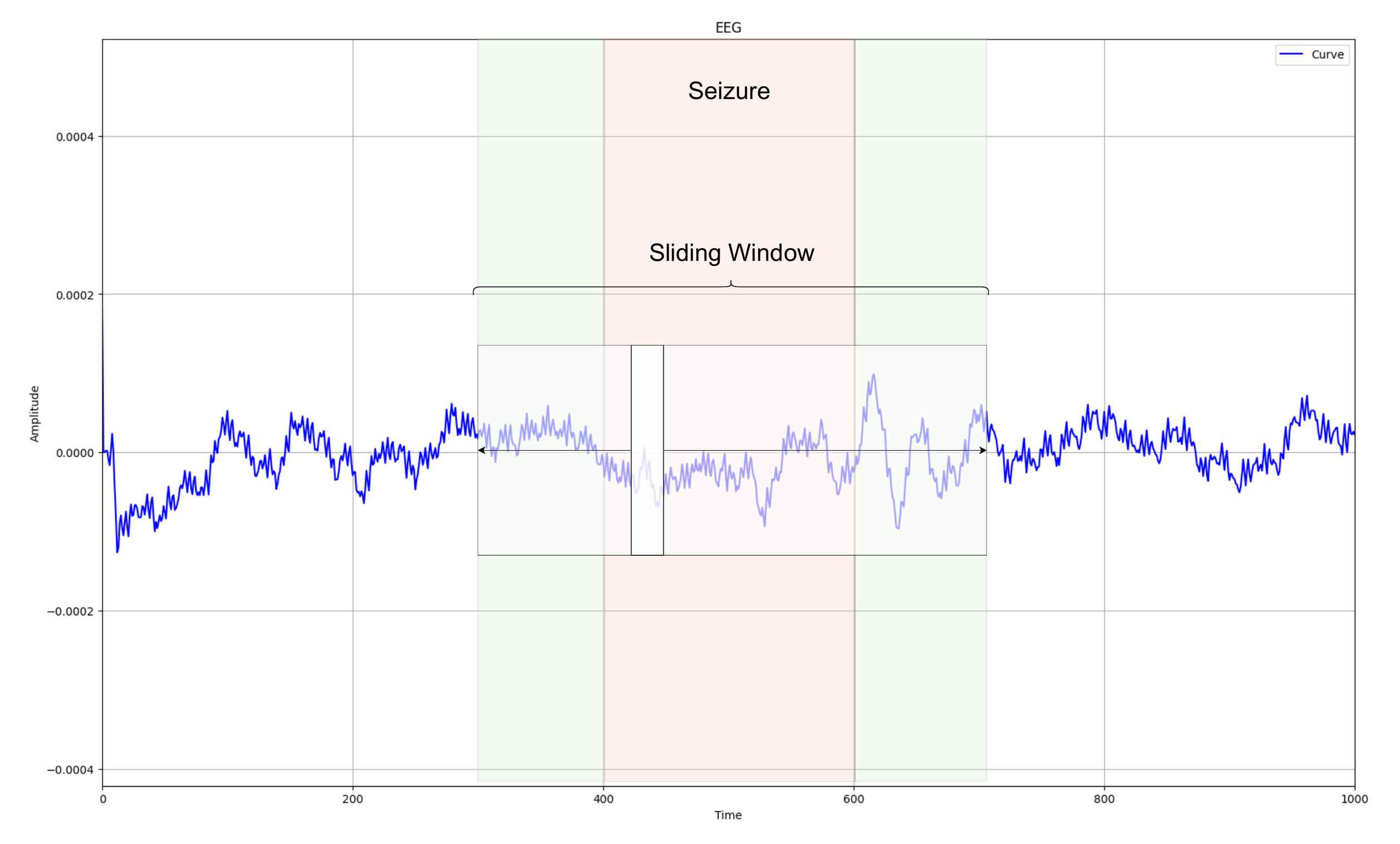}
	\caption{Example of the segmentation process centered in seizures.}
	\label{fig:segmentation}
\end{figure}

Afterward, EEG records are segmented into time windows to address varying seizure durations while ensuring a uniform sample size. These windows are centered around seizure events to improve detection efficiency and reduce computational overhead compared to processing entire recordings. Figure \ref{fig:segmentation} illustrates this segmentation process, where the red area represents seizure activity, and the green area corresponds to non-seizure periods. The number of data points within each seizure segment is proportional to the seizure duration, while each non-seizure segment contains half as many points to maintain balance between seizure and non-seizure instances.

The number of samples generated depends on the window length used during preprocessing. In this study, we applied windows of 1, 2, 5, and 10 seconds. To explore how window duration affects model classification. In time-domain representations, larger windows result in more data points per window, while in frequency-domain representations, larger windows provide better frequency resolution. For instance, a 1-second window results in frequency bins from 0 Hz to 128 Hz with a resolution of 1 Hz, whereas a 10-second window improves resolution to 0.1 Hz.

To increase the amount of training data, we employ overlapping windows as a data augmentation strategy. Each window is shifted by 1 second, resulting in overlaps of 0, 1, 4, and 9 seconds for window sizes of 1, 2, 5, and 10 seconds, respectively. This augmentation yields a total of 17860, 17250, 15440, and 12740 samples.

For windows greater than 10 seconds, the segmentation process would increasingly capture entire seizure episodes within a single window. This reduce the model’s ability to learn seizure onset and offset characteristics. Additionally, larger windows would lead to a decrease in the number of available training samples, limiting the model’s exposure to diverse seizure patterns and potentially impacting generalization. Therefore, we restrict the maximum window size to 10 seconds.

\subsubsection{Data Representations}

Aligned with the data domains and methods described in Section \ref{section:EEG} and Section \ref{section:spectral} (time, frequency, and time-frequency), we employed distinct approaches to generate different representations:

\begin{itemize}
    \item \textbf{Time:} Raw EEG data is used.
    \item \textbf{Frequency:} PSD generated by the Welch's method.
    \item \textbf{Frequency:} PSD generated by the Multitaper method.
    \item \textbf{Time-frequency:} Spectrogram generated by the Multitaper method.
\end{itemize}

As a result of preprocessing, each sample is represented as a segment in the time, frequency, or time-frequency domain. The dimensional format for each segment is structured as 
$(18,X)$ where $18$ represents the number of EEG channels, and $X$ varies depending on the domain (consider $W_s$ as the sliding window size, and $S_r$ the dataset sampling rate):

\begin{itemize}
    \item \textbf{Time:} $(18,S_r \times W_s)$, where $S_r \times W_s$ represents the number of time bins.
    \item \textbf{Frequency:} $(18, \frac{S_r}{2} \times W_s)$, where $\frac{S_r}{2} \times W_s$ corresponds to the number of frequency bins between $0Hz$ and $\frac{S_r}{2}Hz$.
    \item \textbf{Time-frequency:} $(18, F, S_r \times W_s)$, where $F$ is the number of frequency bins between $0 Hz$ and $60 Hz$, and $S_r \times W_s$ represents the number of time bins.
\end{itemize}

\subsubsection{Normalization}

Subsequently, the min-max scaler from Keras \cite{chollet2015keras} is applied to normalize the data within the range $[0, 1]$.  This process improves the convergence speed of the training algorithm and ensures that no single feature disproportionately influences the learning process \cite{han2022data}.

\subsection{Predictive Models}

In this study, predictive models are built to classify EEG segments into normal and ictal (abnormal) classes. For this, the following artificial neural network architectures are used: Recurrent Neural Networks (RNN), Convolutional Neural Networks (CNN), and Convolutional Recurrent Neural Networks (CRNN). These architectures are chosen due to their ability to handle both spatial and temporal aspects of EEG data. RNNs, especially LSTMs, excel at capturing temporal dependencies in sequential data. CNNs are proficient in learning spatial features from the input, while CRNNs combine both capabilities, making them particularly suited for EEG data, which exhibits both spatial and temporal patterns.

Regardless of its architecture, each model ends with a dense network designed for binary classification. This network comprises dense layers utilizing ReLU activation function to introduce non-linearity and are interspersed with dropout layers to prevent overfitting during training \cite{srivastava2014dropout}. At the end of the dense network is a final dense layer employing a Sigmoid activation function, which is used for binary classification.

The CNN model architecture utilizes convolutional layers, pooling, and dropout layers. Convolutional layers are designed to automatically and adaptively learn spatial hierarchies of features from input \cite{lecun1998gradient}. Pooling layers reduce the spatial dimensions of the data, thereby decreasing the computational cost and controlling overfitting \cite{lecun1998gradient}.

The RNN model architecture considered in this work is the Long Short-Term Memory (LSTM), which are well-suited for capturing temporal dependencies in sequential data and maintaining long-term dependencies \cite{hochreiter1997long}. The LSTM layers in our architectures enable the model to effectively learn from data by selectively retaining relevant information across sequences.

The CRNN model integrates convolutional, pooling, dropout, and LSTM layers to leverage the strengths of both CNNs and RNNs. The convolutional and pooling layers extract spatial features from the data, while the LSTM layers capture temporal dependencies.

\subsubsection{Model Hyperparameter Settings}

All models are generated using the Python Keras \cite{chollet2015keras} library due to its ease of use and extensive resources for implementing artificial neural networks. Hyperparameter tuning is applied to key parameters of the RNN, CNN, CRNN layers, and the dense network layers at the end of each model to ensure optimal training conditions.

This tuning process is performed using Bayesian optimization from Keras Tuner \cite{omalley2019kerastuner}. The search space explores different configurations over 20 trials, using a batch size of 32 (default) and training each candidate model for 50 epochs. After all trials are completed, the best-performing hyperparameter set is selected based on the validation performance.

The RNN hyperparameters are fine-tuned by adjusting the number of LSTM units, dropout rates, and units on dense layers. The RNN model architecture and its chosen hyperparameter space are shown in Table~\ref{tab:rnn_hyperparameter_space}.

\begin{table}[]
\caption{Hyperparameter space for RNN}
\centering
\begin{tabular}{|ccc|}
\hline
\multicolumn{3}{|c|}{\textbf{RNN Model}} \\
\hline
Layer Type & Min Value & Max Value \\ \hline
LSTM       & 8         & 64        \\
Dropout    & 0.1       & 0.8       \\
LSTM       & 8         & 32        \\
Dropout    & 0.1       & 0.8       \\ 
Flatten    & -         & -         \\
Dense      & 32        & 256       \\
Dropout    & 0.1       & 0.8       \\
Dense      & 16        & 128       \\
Dropout    & 0.1       & 0.8       \\
Dense      & 1         & 1         \\ \hline
\end{tabular}%
\label{tab:rnn_hyperparameter_space}
\end{table}

For CNN models, the adjusted hyperparameters include the number of convolutional filters, dropout rates, and units on dense layers. The CNN model architecture and its chosen hyperparameter space are shown in Table \ref{tab:cnn_hyperparameter_space}.

\begin{table}[]
\caption{Hyperparameter space for CNN}
\centering
\begin{tabular}{|ccc|}
\hline
\multicolumn{3}{|c|}{\textbf{CNN Model}} \\
\hline
Layer Type  & Min Value & Max Value \\ \hline
Convolution & 8         & 64        \\
MaxPooling  & (3,3)     & (3,3)     \\
Dropout     & 0.1       & 0.8       \\
Convolution & 16        & 128       \\
MaxPooling  & (3,3) || (2,2)     & (3,3) || (2,2)     \\
Dropout     & 0.1       & 0.8       \\ 
Flatten     & -       & -       \\
Dense      & 32        & 256       \\
Dropout    & 0.1       & 0.8       \\
Dense      & 16        & 128       \\
Dropout    & 0.1       & 0.8       \\
Dense      & 1         & 1         \\ \hline
\end{tabular}%
\label{tab:cnn_hyperparameter_space}
\end{table}

For CRNN models, the adjusted hyperparameters include the number of convolutional filters, dropout rates, LSTM units, and units on dense layers. The CNN model architecture and its hyperparameter space are shown in Table \ref{tab:crnn_hyperparameter_space}.

\begin{table}[]
\caption{Hyperparameter space for CRNN}
\centering
\begin{tabular}{|ccc|}
\hline
\multicolumn{3}{|c|}{\textbf{CRNN Model}} \\
\hline
Layer Type  & Min Value & Max Value \\ \hline
Convolution & 8         & 64        \\
MaxPooling  & (3,3)     & (3,3)     \\
Dropout     & 0.1       & 0.8       \\
Convolution & 16        & 128       \\
MaxPooling  & (3,3) || (2,2)     & (3,3) || (2,2)     \\
Dropout     & 0.1       & 0.8       \\
Flatten     & -       & -       \\
LSTM        & 8         & 64        \\
LSTM        & 8         & 32        \\
Dense      & 32        & 256       \\
Dropout    & 0.1       & 0.8       \\
Dense      & 16        & 128       \\
Dropout    & 0.1       & 0.8       \\
Dense      & 1         & 1         \\ \hline
\end{tabular}%
\label{tab:crnn_hyperparameter_space}
\end{table}

The learning rate is also optimized using the Keras Tuner hyperparameter tuning, with a value between $1\times 10^{-5}$, and $1\times10^{-3}$. This range was selected to balance training stability and convergence speed, allowing the models to learn efficiently without overshooting or slow progress.

\subsubsection{Training Models}

For the final training, the batch size was set to 16, and the number of epochs was doubled from the tuner setting, reaching 100 epochs. This decision aimed to provide more training cycles for further model refinement.

A custom model checkpoint callback is created during training to monitor validation accuracy. The checkpoint saved the model weights with the best validation performance, ensuring that the most effective model configuration was captured without overfitting the training data.

The cross-validation method employed in this work is the k-fold cross-validation \cite{refaeilzadeh2009cross}, where we have implemented a 10-fold cross-validation.

A total of 48 models were built considering three different model architectures, four data representation, and four sliding window sizes. Each model architecture is trained separately across each combination of data representation and sliding window size to assess its effectiveness under different conditions.

\subsubsection{Data Splitting}

To ensure robust model evaluation and prevent information leakage, the dataset is partitioned into distinct subsets for hyperparameter tuning, model training and testing.

First, 20\% of the dataset is set aside exclusively for hyperparameter tuning. This subset is further split into two parts: 80\% is used to train models with different hyperparameter configurations, while the remaining 20\% serves as a validation set to assess their performance and guide the selection of the optimal hyperparameters.

The remaining 80\% of the dataset is used for model training, validation, and final testing. This portion is further divided into 64\% for model training, 16\% for validation, and the final 20\% for testing. The test set remains completely untouched during both the hyperparameter tuning and model training phases, granting the objective to test on unseen data.

\subsection{Model Evaluation}
Confusion matrix metrics are commonly used in the literature to evaluate the models. This matrix is composed of the following parameters \cite{sokolova2009systematic}:

\begin{itemize}
    \item \textbf{True Positives} (TP): Number of abnormal segments (with anomaly) correctly classified.
    \item \textbf{True Negatives} (TN): Number of normal segments correctly classified.
    \item \textbf{False Negatives} (FN): Number of abnormal segments incorrectly classified.
    \item \textbf{False Positives} (FP): Number of normal segments incorrectly classified.
\end{itemize}

Using these parameters from the confusion matrix, the following metrics were calculated:

\begin{itemize}
    \item \textbf{Accuracy} (Equation \ref{eq:Accuracy}): The overall rate of correctly classified segments.
    \begin{equation}
        \label{eq:Accuracy}
        Accuracy = \frac{TP + TN}{TP + TN + FP + FN}
        \end{equation}
        
    \item \textbf{Precision} (Equation \ref{eq:Precision}): For all segments classified as abnormal, this measure determines the percentage that are correctly categorized.
    \begin{equation}
        \label{eq:Precision}
        Precision = \frac{TP}{TP + FP}
    \end{equation}

    \item \textbf{Sensitivity} (Equation \ref{eq:Sensitivity}): Also known as recall, it evaluates the model's ability to successfully detect positive results.
    \begin{equation}
        \label{eq:Sensitivity}
        Sensitivity = \frac{TP}{TP + FN}
    \end{equation}

    \item \textbf{Specificity} (Equation \ref{eq:Specificity}): Evaluates the model's ability to detect negative results.
    \begin{equation}
        \label{eq:Specificity}
        Specificity = \frac{TN}{TN + FP}
    \end{equation}

    \item \textbf{F1-Score} (Equation \ref{eq:F1-Score}): A harmonic mean calculated based on precision and sensitivity.
    \begin{equation}
        \label{eq:F1-Score}
        F1-Score = \frac{2*Precision*Sensitivity}{Precision + Sensitivity}
    \end{equation}
\end{itemize}

In addition to the above metrics, Receiver Operating Characteristic (ROC) curves and the Area Under the Curve (AUC) are also used in the model evaluation to demonstrate the performance of a learning model through the relationship between the true positive rate and the false positive rate. The AUC provides an aggregate measure of performance across all classification thresholds.

To complement these metrics, a statistical hypothesis test was conducted for unpaired and non-parametric data. The Kruskal-Wallis \cite{kruskal1952use} test was applied to determine whether there was a statistically significant difference between the models. Since it identifies if there are differences between groups but does not specify which pairs of groups are different, a post-hoc test is performed to further analyze the results. The Dunn \cite{dunn1961multiple} test was used for pairwise comparisons, allowing us to identify which specific groups show statistically significant differences.

The Kruskal-Wallis test was applied to groups defined by different window sizes and representations, as these factors represent independent experimental conditions.

For non-parametric data with repeated measures, the Friedman test \cite{friedman1937use} was applied to determine whether there were significant differences between the models across different experimental conditions. Unlike the Kruskal-Wallis test, which is used for unpaired data, the Friedman test is specifically designed for related or paired samples, making it suitable for scenarios where the same subjects or experimental units are measured multiple times. Since the Friedman test only identifies whether there is a significant difference between groups, but does not specify which groups differ from each other, a post-hoc analysis was performed. The Nemenyi post-hoc test \cite{nemenyi1963distribution} was used to conduct pairwise comparisons between the models, identifying which specific groups showed statistically significant differences.

The Friedman test was applied to the model architecture group, as this refers to different models tested under the same experimental conditions.

\section{Results and Discussion}\label{section:RESULTS}

The proposed methods and models was developed and tested on a Windows 11 machine equipped with an Intel® Core™ i7-11800H CPU (2.30GHz), 16 GB of RAM, and an NVIDIA GeForce RTX 3060 Laptop GPU.
Additionally, some experiments were conducted using Google Colaboratory, which provided access to 53 GB of RAM and an NVIDIA L4 GPU. The implementation was carried out using Python for both data preprocessing and the training of deep learning models.

Tables \ref{tab:rnn_performance}, \ref{tab:cnn_performance}, and \ref{tab:crnn_performance} summarize the performance of the RNN, CNN, and CRNN models, respectively, across all different data domains and sliding windows. Gray values indicate the highest value for each metric and window.

\begin{table}[]
\caption{Evaluation of RNN}
\centering
\resizebox{\textwidth}{!}{
\begin{tabular}{|c|c|c|c|c|c|}
\hline
\multicolumn{6}{|c|}{\textbf{RNN Model}} \\ \hline
\textbf{Metric} & \textbf{Window} & \textbf{Time} & \textbf{Welch} & \textbf{Multitaper} & \textbf{Spectrogram} \\ \hline
\multirow{4}{*}{\textbf{Accuracy}} 
    & 1s  & $78.49\pm2.43$ & $84.78\pm0.46$ & \cellcolor{lightgray}$87.86\pm0.49$ & $83.47\pm1.48$ \\
    & 2s  & $79.74\pm1.66$ & $85.40\pm0.37$ & \cellcolor{lightgray}$91.65\pm0.56$ & $83.73\pm1.71$ \\
    & 5s  & $77.29\pm0.84$ & $92.96\pm0.49$ & \cellcolor{lightgray}$95.99\pm0.19$ & $80.91\pm1.10$ \\
    & 10s & $76.83\pm1.78$ & $96.18\pm0.43$ & \cellcolor{lightgray}$97.05\pm0.24$ & $79.67\pm2.08$ \\ \hline

\multirow{4}{*}{\textbf{Precision}} 
    & 1s  & $79.26\pm3.30$ & $85.64\pm2.02$ & \cellcolor{lightgray}$89.45\pm1.54$ & $86.61\pm2.90$ \\
    & 2s  & $80.45\pm4.41$ & $87.21\pm1.20$ & \cellcolor{lightgray}$93.39\pm1.18$ & $85.09\pm2.15$ \\
    & 5s  & $75.06\pm4.41$ & $94.09\pm1.22$ & \cellcolor{lightgray}$96.59\pm0.58$ & $82.08\pm3.30$ \\
    & 10s & $78.58\pm4.42$ & $97.37\pm0.59$ & \cellcolor{lightgray}$97.55\pm0.53$ & $81.53\pm3.89$ \\ \hline

\multirow{4}{*}{\textbf{Sensitivity}} 
    & 1s  & $78.10\pm2.92$ & $84.07\pm1.42$ & \cellcolor{lightgray}$86.58\pm1.17$ & $81.37\pm1.79$ \\
    & 2s  & $79.11\pm2.30$ & $84.20\pm1.15$ & \cellcolor{lightgray}$90.28\pm1.22$ & $82.88\pm2.10$ \\
    & 5s  & $78.98\pm3.05$ & $92.04\pm1.31$ & \cellcolor{lightgray}$95.44\pm0.70$ & $80.19\pm2.57$ \\
    & 10s & $75.50\pm3.16$ & $94.96\pm0.77$ & \cellcolor{lightgray}$96.47\pm0.52$ & $78.17\pm3.89$ \\ \hline

\multirow{4}{*}{\textbf{Specificity}} 
    & 1s  & $79.05\pm2.78$ & $85.61\pm1.50$ & \cellcolor{lightgray}$89.27\pm1.27$ & $85.97\pm2.49$ \\
    & 2s  & $80.69\pm3.06$ & $86.75\pm0.87$ & \cellcolor{lightgray}$93.17\pm1.06$ & $84.70\pm1.99$ \\
    & 5s  & $76.17\pm2.34$ & $93.99\pm1.11$ & \cellcolor{lightgray}$96.56\pm0.54$ & $81.95\pm2.20$ \\
    & 10s & $78.73\pm2.56$ & $97.42\pm0.56$ & \cellcolor{lightgray}$97.62\pm0.50$ & $81.74\pm2.41$ \\ \hline

\multirow{4}{*}{\textbf{F1-Score}} 
    & 1s  & $78.29\pm2.59$ & $84.41\pm0.83$ & \cellcolor{lightgray}$87.21\pm0.74$ & $82.40\pm1.50$ \\
    & 2s  & $79.41\pm1.73$ & $84.79\pm0.72$ & \cellcolor{lightgray}$90.96\pm0.83$ & $83.30\pm1.85$ \\
    & 5s  & $78.10\pm1.72$ & $92.50\pm0.85$ & \cellcolor{lightgray}$95.71\pm0.42$ & $80.53\pm1.70$ \\
    & 10s & $76.15\pm2.38$ & $95.56\pm0.57$ & \cellcolor{lightgray}$96.76\pm0.33$ & $78.89\pm2.08$ \\ \hline
\end{tabular}%
}
\label{tab:rnn_performance}
\end{table}

\begin{table}[]
\caption{Evaluation of CNN}
\centering
\resizebox{\textwidth}{!}{
\begin{tabular}{|c|c|c|c|c|c|}
\hline
\multicolumn{6}{|c|}{\textbf{CNN Model}} \\ \hline
\textbf{Metric} & \textbf{Window} & \textbf{Time} & \textbf{Welch} & \textbf{Multitaper} & \textbf{Spectrogram} \\ \hline
\multirow{4}{*}{\textbf{Accuracy}} 
    & 1s  & $82.01\pm0.81$ & $84.70\pm0.32$ & \cellcolor{lightgray}$88.51\pm0.24$ & $84.04\pm0.57$ \\
    & 2s  & $83.59\pm0.55$ & $88.95\pm0.37$ & \cellcolor{lightgray}$93.61\pm0.32$ & $89.20\pm0.35$ \\
    & 5s  & $86.71\pm0.47$ & $94.07\pm0.26$ & \cellcolor{lightgray}$96.82\pm0.24$ & $86.34\pm0.64$ \\
    & 10s & $83.19\pm0.58$ & $96.03\pm0.24$ & \cellcolor{lightgray}$97.63\pm0.16$ & $87.12\pm1.84$ \\ \hline

\multirow{4}{*}{\textbf{Precision}} 
    & 1s  & $84.39\pm1.84$ & $86.93\pm0.87$ & \cellcolor{lightgray}$89.76\pm1.57$ & $87.71\pm1.92$ \\
    & 2s  & $88.28\pm2.64$ & $91.34\pm0.80$ & \cellcolor{lightgray}$93.65\pm0.77$ & $91.54\pm1.08$ \\
    & 5s  & $91.11\pm1.22$ & $95.24\pm0.77$ & \cellcolor{lightgray}$97.91\pm0.36$ & $93.02\pm1.92$ \\
    & 10s & $84.93\pm2.24$ & $97.43\pm0.42$ & \cellcolor{lightgray}$97.85\pm0.36$ & $91.62\pm5.69$ \\ \hline

\multirow{4}{*}{\textbf{Sensitivity}} 
    & 1s  & $80.44\pm1.71$ & $83.06\pm0.59$ & \cellcolor{lightgray}$87.47\pm1.32$ & $81.62\pm1.75$ \\
    & 2s  & $80.77\pm1.42$ & $87.19\pm0.74$ & \cellcolor{lightgray}$93.58\pm0.60$ & $87.46\pm1.04$ \\
    & 5s  & $83.76\pm1.15$ & $93.07\pm0.56$ & \cellcolor{lightgray}$95.82\pm0.55$ & $82.02\pm1.70$ \\
    & 10s & $81.61\pm1.92$ & $94.62\pm0.34$ & \cellcolor{lightgray}$97.33\pm0.43$ & $83.84\pm3.30$ \\ \hline

\multirow{4}{*}{\textbf{Specificity}} 
    & 1s  & $83.84\pm1.26$ & $86.49\pm0.70$ & \cellcolor{lightgray}$89.66\pm1.24$ & $86.96\pm1.48$ \\
    & 2s  & $87.18\pm2.15$ & $90.91\pm0.70$ & \cellcolor{lightgray}$93.65\pm0.69$ & $91.15\pm0.92$ \\
    & 5s  & $90.29\pm1.06$ & $95.14\pm0.72$ & \cellcolor{lightgray}$97.87\pm0.35$ & $92.12\pm1.79$ \\
    & 10s & $85.00\pm1.50$ & $97.46\pm0.40$ & \cellcolor{lightgray}$97.93\pm0.34$ & $91.70\pm4.79$ \\ \hline

\multirow{4}{*}{\textbf{F1-Score}} 
    & 1s  & $81.21\pm1.20$ & $83.87\pm0.40$ & \cellcolor{lightgray}$87.98\pm0.74$ & $82.81\pm1.15$ \\
    & 2s  & $82.15\pm0.82$ & $88.06\pm0.52$ & \cellcolor{lightgray}$93.59\pm0.39$ & $88.32\pm0.66$ \\
    & 5s  & $85.21\pm0.78$ & $93.56\pm0.33$ & \cellcolor{lightgray}$96.32\pm0.38$ & $84.12\pm1.15$ \\
    & 10s & $82.38\pm1.21$ & $95.32\pm0.26$ & \cellcolor{lightgray}$97.48\pm0.27$ & $85.41\pm2.19$ \\ \hline
\end{tabular}%
}
\label{tab:cnn_performance}
\end{table}

\begin{table}[]
\caption{Evaluation of CRNN}
\centering
\resizebox{\textwidth}{!}{
\begin{tabular}{|c|c|c|c|c|c|}
\hline
\multicolumn{6}{|c|}{\textbf{CRNN Model}} \\ \hline
\textbf{Metric} & \textbf{Window} & \textbf{Time} & \textbf{Welch} & \textbf{Multitaper} & \textbf{Spectrogram} \\ \hline
\multirow{4}{*}{\textbf{Accuracy}} 
    & 1s  & $81.93\pm0.55$ & $84.73\pm0.36$ & \cellcolor{lightgray}$88.21\pm0.32$ & $84.11\pm0.74$ \\
    & 2s  & $85.63\pm0.57$ & $87.01\pm0.48$ & \cellcolor{lightgray}$92.38\pm0.37$ & $88.28\pm0.69$ \\
    & 5s  & $88.27\pm1.48$ & $93.38\pm0.25$ & \cellcolor{lightgray}$95.88\pm0.32$ & $86.11\pm0.66$ \\
    & 10s & $90.08\pm1.64$ & $95.54\pm0.49$ & \cellcolor{lightgray}$96.87\pm0.32$ & $86.74\pm1.56$ \\ \hline

\multirow{4}{*}{\textbf{Precision}} 
    & 1s  & $86.66\pm2.57$ & $86.27\pm1.45$ & \cellcolor{lightgray}$89.94\pm1.65$ & $91.05\pm1.29$ \\
    & 2s  & $88.19\pm3.49$ & $88.48\pm0.93$ & \cellcolor{lightgray}$93.15\pm0.84$ & $91.65\pm1.90$ \\
    & 5s  & $91.69\pm1.31$ & $93.60\pm1.16$ & \cellcolor{lightgray}$97.13\pm0.67$ & $88.37\pm3.14$ \\
    & 10s & $93.43\pm3.41$ & $96.07\pm0.86$ & \cellcolor{lightgray}$97.98\pm0.47$ & $88.21\pm4.31$ \\ \hline

\multirow{4}{*}{\textbf{Sensitivity}} 
    & 1s  & $79.09\pm1.98$ & $83.57\pm1.26$ & \cellcolor{lightgray}$86.84\pm1.12$ & $79.84\pm1.41$ \\
    & 2s  & $84.01\pm1.86$ & $85.97\pm1.02$ & \cellcolor{lightgray}$91.75\pm0.69$ & $85.91\pm1.53$ \\
    & 5s  & $85.88\pm2.27$ & $93.21\pm0.97$ & \cellcolor{lightgray}$94.76\pm0.45$ & $84.82\pm1.49$ \\
    & 10s & $87.28\pm1.90$ & $94.90\pm0.85$ & \cellcolor{lightgray}$95.74\pm0.73$ & $85.27\pm2.23$ \\ \hline

\multirow{4}{*}{\textbf{Specificity}} 
    & 1s  & $85.57\pm1.91$ & $86.03\pm1.00$ & \cellcolor{lightgray}$89.75\pm1.35$ & $89.78\pm1.13$ \\
    & 2s  & $87.73\pm2.69$ & $88.14\pm0.74$ & \cellcolor{lightgray}$93.05\pm0.75$ & $91.11\pm1.66$ \\
    & 5s  & $91.10\pm1.26$ & $93.61\pm1.04$ & \cellcolor{lightgray}$97.07\pm0.65$ & $87.78\pm2.58$ \\
    & 10s & $93.37\pm3.14$ & $96.19\pm0.79$ & \cellcolor{lightgray}$98.02\pm0.44$ & $88.59\pm3.44$ \\ \hline

\multirow{4}{*}{\textbf{F1-Score}} 
    & 1s  & $80.48\pm1.26$ & $84.14\pm0.76$ & \cellcolor{lightgray}$87.52\pm0.61$ & $81.92\pm1.07$ \\
    & 2s  & $84.80\pm0.86$ & $86.49\pm0.71$ & \cellcolor{lightgray}$92.06\pm0.47$ & $87.08\pm1.02$ \\
    & 5s  & $87.06\pm1.85$ & $93.29\pm0.53$ & \cellcolor{lightgray}$95.32\pm0.32$ & $85.45\pm0.69$ \\
    & 10s & $88.65\pm1.55$ & $95.22\pm0.61$ & \cellcolor{lightgray}$96.30\pm0.50$ & $85.98\pm1.56$ \\ \hline
\end{tabular}%
}
\label{tab:crnn_performance}
\end{table}

To provide a comprehensive summary of the model performance across different evaluation metrics, Table \ref{tab:model_rank_by_metrics} presents the top ten models for each metric, ranked by their mean values.

\begin{table}[]
\caption{Rank of models by metric}
\centering
\resizebox{\textwidth}{!}{%
\begin{tabular}{|ccccc|}
\hline
 &  & Rank &  &  \\ \hline
\multicolumn{1}{|c|}{Accuracy} & \multicolumn{1}{c|}{Precision} & \multicolumn{1}{c|}{Sensitivity} & \multicolumn{1}{c|}{Specificity} & F1-Score \\ \hline
\multicolumn{1}{|c|}{CNN - Multitaper - 10s} & \multicolumn{1}{c|}{CRNN - Multitaper - 10s} & \multicolumn{1}{c|}{CNN - Multitaper - 10s} & \multicolumn{1}{c|}{CRNN - Multitaper - 10s} & CNN - Multitaper - 10s \\
\multicolumn{1}{|c|}{RNN - Multitaper - 10s} & \multicolumn{1}{c|}{CNN - Multitaper - 5s} & \multicolumn{1}{c|}{RNN - Multitaper - 10s} & \multicolumn{1}{c|}{CNN - Multitaper - 10s} & RNN - Multitaper - 10s \\
\multicolumn{1}{|c|}{CRNN - Multitaper - 10s} & \multicolumn{1}{c|}{CNN - Multitaper - 10s} & \multicolumn{1}{c|}{CNN - Multitaper - 5s} & \multicolumn{1}{c|}{CNN - Multitaper - 5s} & CNN - Multitaper - 5s \\
\multicolumn{1}{|c|}{CNN - Multitaper - 5s} & \multicolumn{1}{c|}{RNN - Multitaper - 10s} & \multicolumn{1}{c|}{CRNN - Multitaper - 10s} & \multicolumn{1}{c|}{RNN - Multitaper - 10s} & CRNN - Multitaper - 10s \\
\multicolumn{1}{|c|}{RNN - Welch - 10s} & \multicolumn{1}{c|}{CNN - Welch - 10s} & \multicolumn{1}{c|}{RNN - Multitaper - 5s} & \multicolumn{1}{c|}{CNN - Welch - 10s} & RNN - Welch - 10s \\
\multicolumn{1}{|c|}{CNN - Welch - 10s} & \multicolumn{1}{c|}{RNN - Welch - 10s} & \multicolumn{1}{c|}{RNN - Welch - 10s} & \multicolumn{1}{c|}{RNN - Welch - 10s} & CNN - Welch - 10s \\
\multicolumn{1}{|c|}{RNN - Multitaper - 5s} & \multicolumn{1}{c|}{CRNN - Multitaper - 5s} & \multicolumn{1}{c|}{CRNN - Welch - 10s} & \multicolumn{1}{c|}{CRNN - Multitaper - 5s} & CRNN - Multitaper - 5s \\
\multicolumn{1}{|c|}{CRNN - Multitaper - 5s} & \multicolumn{1}{c|}{RNN - Multitaper - 5s} & \multicolumn{1}{c|}{CRNN - Multitaper - 5s} & \multicolumn{1}{c|}{RNN - Multitaper - 5s} & CRNN - Welch - 10s \\
\multicolumn{1}{|c|}{CRNN - Welch - 10s} & \multicolumn{1}{c|}{CRNN - Welch - 10s} & \multicolumn{1}{c|}{CNN - Welch - 10s} & \multicolumn{1}{c|}{CRNN - Welch - 10s} & \multicolumn{1}{l|}{CNN - Multitaper - 2s} \\
\multicolumn{1}{|c|}{CNN - Welch - 5s} & \multicolumn{1}{c|}{CNN - Welch - 5s} & \multicolumn{1}{l|}{CNN - Multitaper - 2s} & \multicolumn{1}{c|}{CNN - Welch - 5s} & CNN - Welch - 5s \\ \hline
\end{tabular}%
}
\label{tab:model_rank_by_metrics}
\end{table}

From the tables, we observe that the Multitaper method outperforms all other representations across all metrics, window sizes, and models. The Welch and Spectrogram methods show fluctuations in performance, especially for smaller window sizes. However, as the window size increases, Welch surpasses the Spectrogram method. These results suggest that the highest-performing models across all metrics were those trained with frequency-domain representations. This indicates that these models possibly leverage specific frequency bands for epileptic seizure detection, as discussed in Section \ref{section:spectral}, where certain frequency bands are associated with neurological disorders.

Although the Multitaper method is the dominant approach, the Welch method demonstrates competitive performance in some metrics. For example, CNN, RNN and CRNN models with Welch-based 10-second windows achieve similar results as the Multitaper-based.

Additionally, increasing the window size improved performance for frequency-domain representations, likely due to the enhanced frequency resolution, as discussed in Subsection \ref{sub:sliding_window}, where longer windows provide higher frequency bin resolution. In the time-domain representation, performance improvements were observed only in the CRNN architecture, with an increase of over 8\% in the metrics. However, for time-frequency representations, no significant improvements were observed.

For time-domain data, we observed the lowest performance metrics among RNN and CNN models. However, in the CRNN model, some metrics exceeded 90\%, suggesting that time-domain data may require further preprocessing to achieve better results, such as applying filters for artifact removal, as mentioned in related literature (Section \ref{section:related}).

\subsection{ROC and AUC Analysis}

For the ROC and AUC analysis, we selected the best-performing model for each combination and evaluated it on 20 dataset files.

\begin{figure}[htbp]
    \centering
    \begin{subfigure}{0.49\textwidth}
        \centering
        \includegraphics[width=\linewidth]{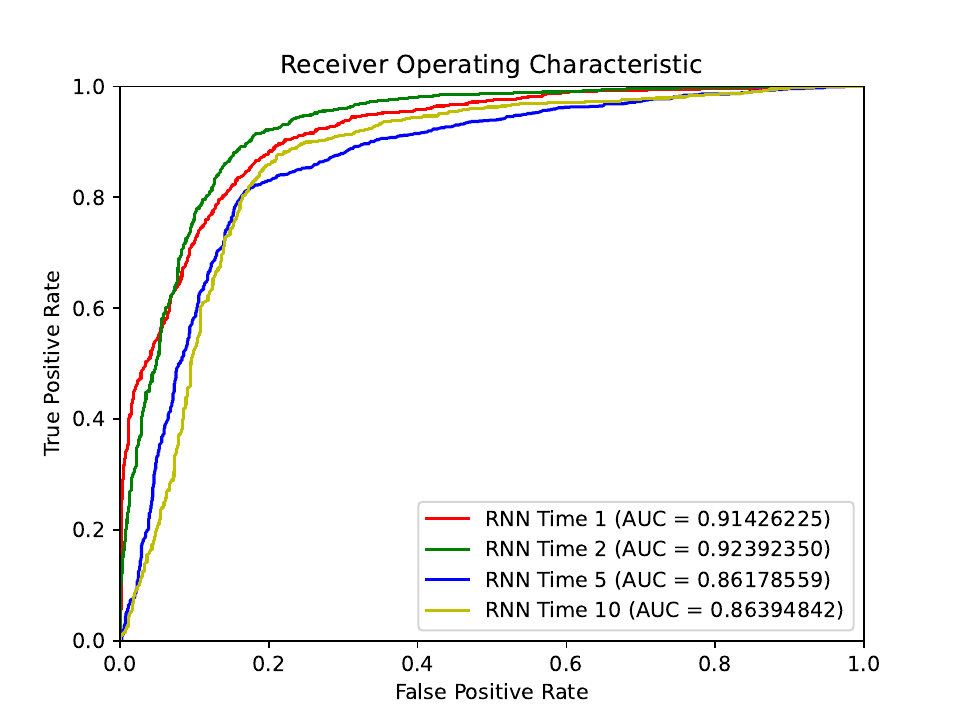}
        \caption{ROC and AUC RNN time}
        \label{fig:rnn_roc_time}
    \end{subfigure}
    \hfill
    \begin{subfigure}{0.49\textwidth}
        \centering
        \includegraphics[width=\linewidth]{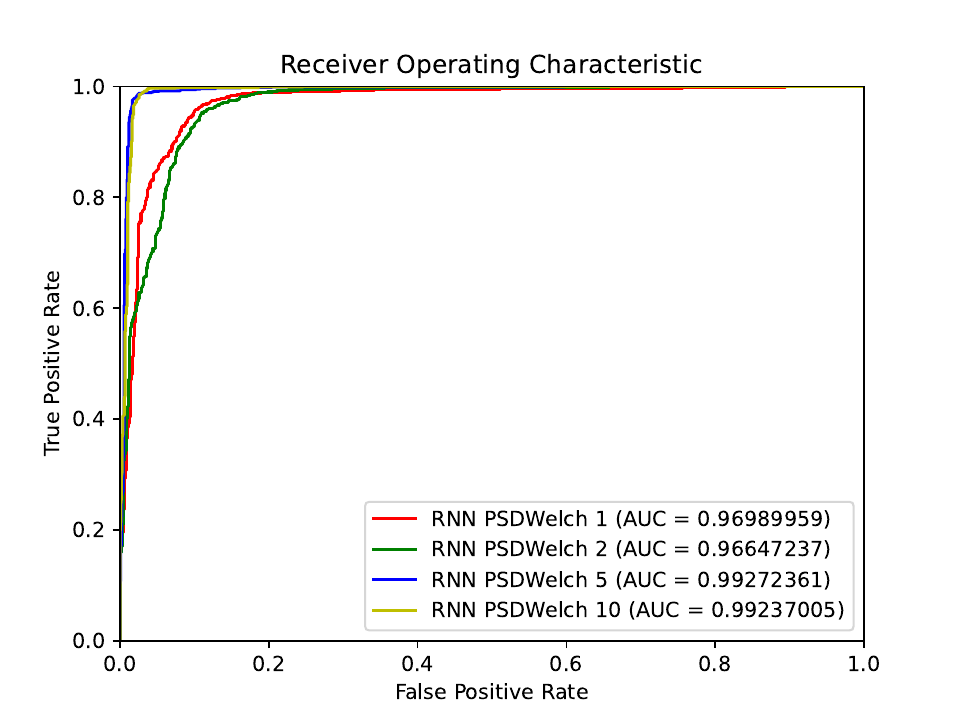}
        \caption{ROC and AUC RNN Welch}
        \label{fig:rnn_roc_welch}
    \end{subfigure}
    \begin{subfigure}{0.49\textwidth}
        \centering
        \includegraphics[width=\linewidth]{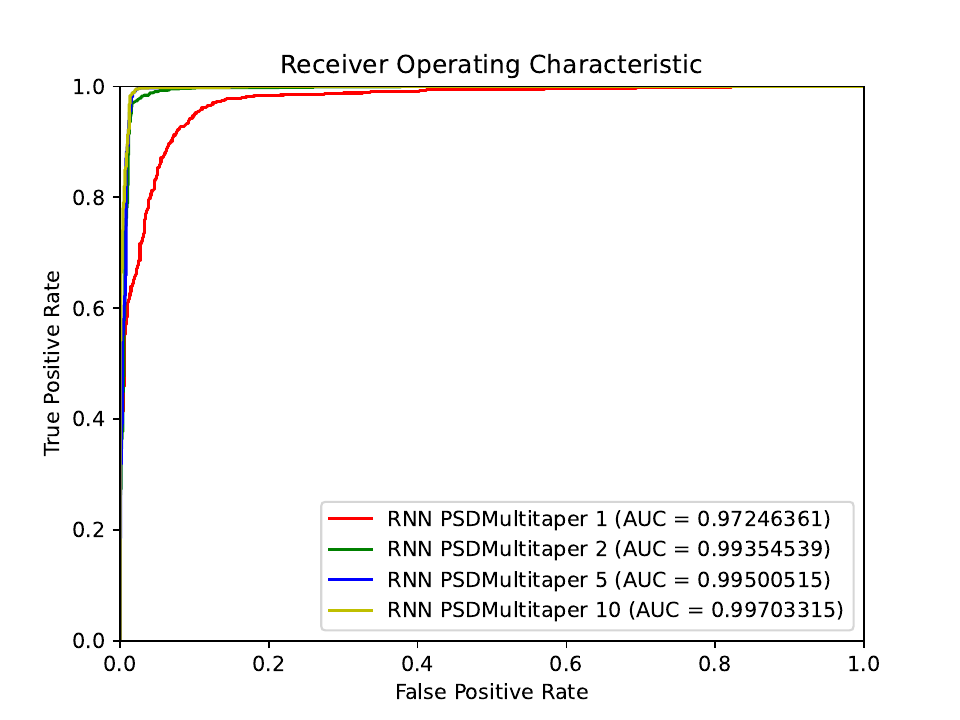}
        \caption{ROC and AUC RNN Multitaper}
        \label{fig:rnn_roc_multi}
    \end{subfigure}
    \hfill
    \begin{subfigure}{0.49\textwidth}
        \centering
        \includegraphics[width=\linewidth]{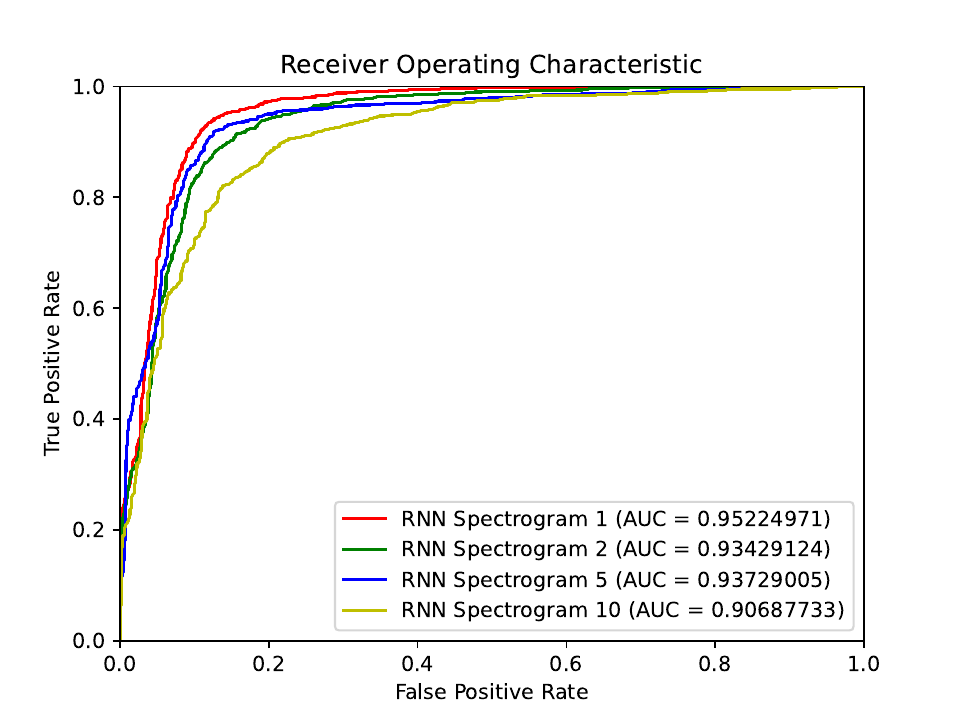}
        \caption{ROC and AUC RNN spectrogram}
        \label{fig:rnn_roc_spec}
    \end{subfigure}
    \caption{ROC and AUC for RNN models}
    \label{fig:rnn_roc_auc}
\end{figure}

For RNN models, Figure \ref{fig:rnn_roc_time} presents the ROC curve and AUC for the time-domain representation, Figure \ref{fig:rnn_roc_welch} for the Welch method, Figure \ref{fig:rnn_roc_multi} for the Multitaper method, and Figure \ref{fig:rnn_roc_spec} for the spectrogram with Multitaper.

\begin{figure}[htbp]
    \centering
    \begin{subfigure}{0.49\textwidth}
        \centering
        \includegraphics[width=\linewidth]{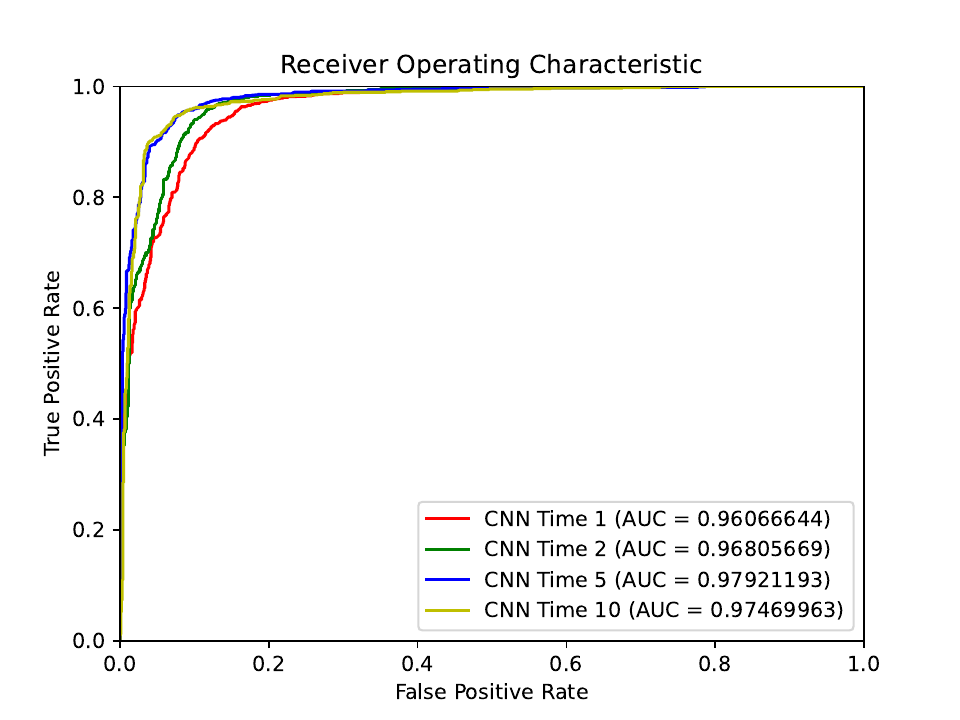}
        \caption{ROC and AUC CNN time}
        \label{fig:cnn_roc_time}
    \end{subfigure}
    \hfill
    \begin{subfigure}{0.49\textwidth}
        \centering
        \includegraphics[width=\linewidth]{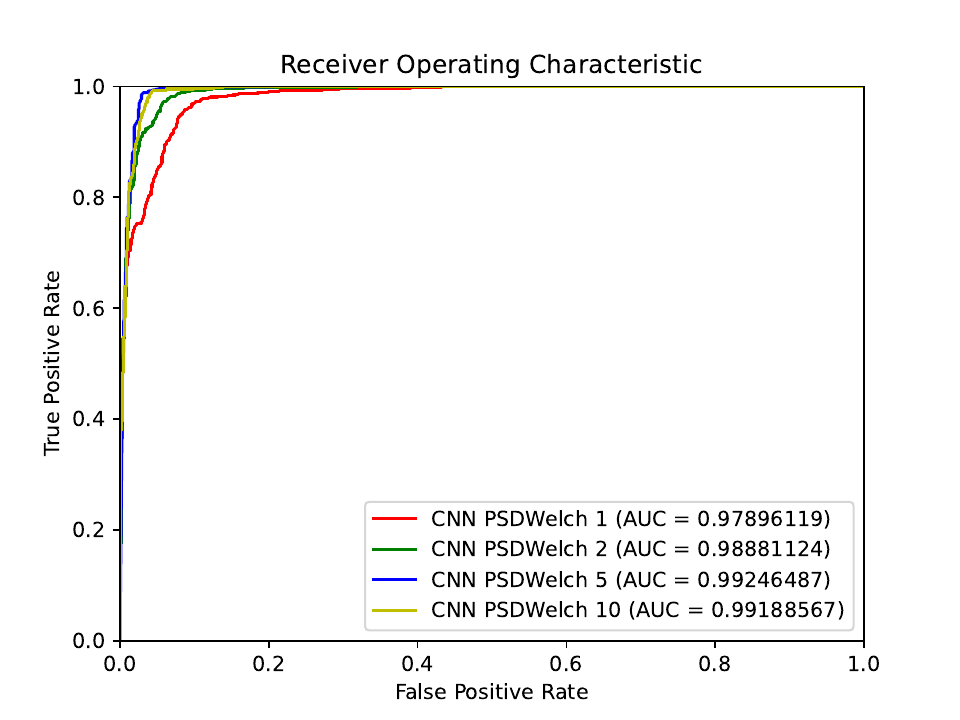}
        \caption{ROC and AUC CNN Welch}
        \label{fig:cnn_roc_welch}
    \end{subfigure}
    \begin{subfigure}{0.49\textwidth}
        \centering
        \includegraphics[width=\linewidth]{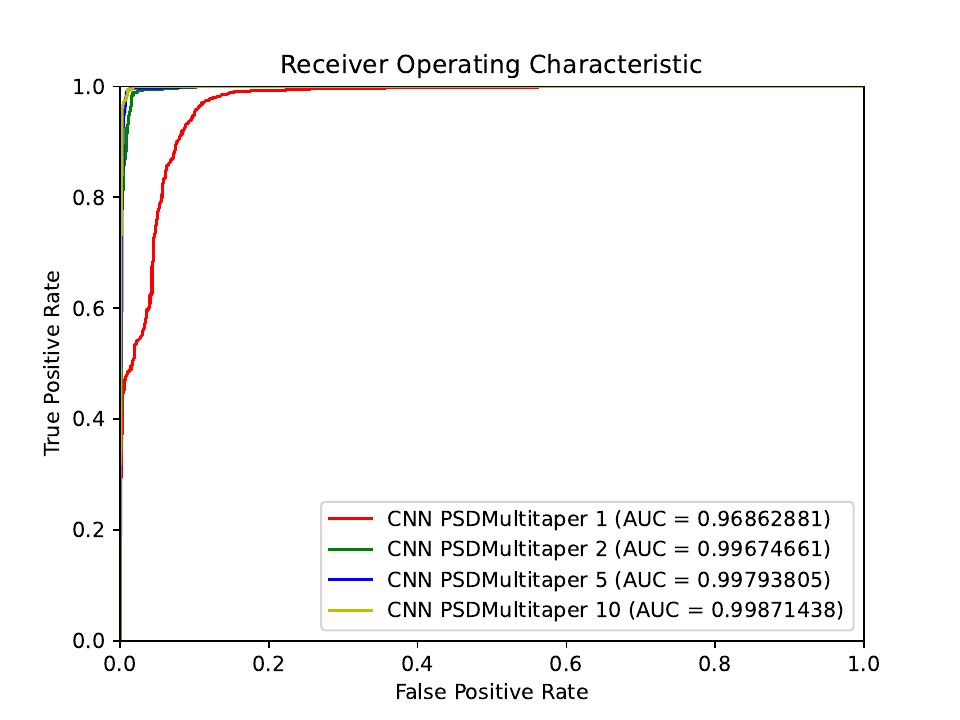}
        \caption{ROC and AUC CNN Multitaper}
        \label{fig:cnn_roc_multi}
    \end{subfigure}
    \hfill
    \begin{subfigure}{0.49\textwidth}
        \centering
        \includegraphics[width=\linewidth]{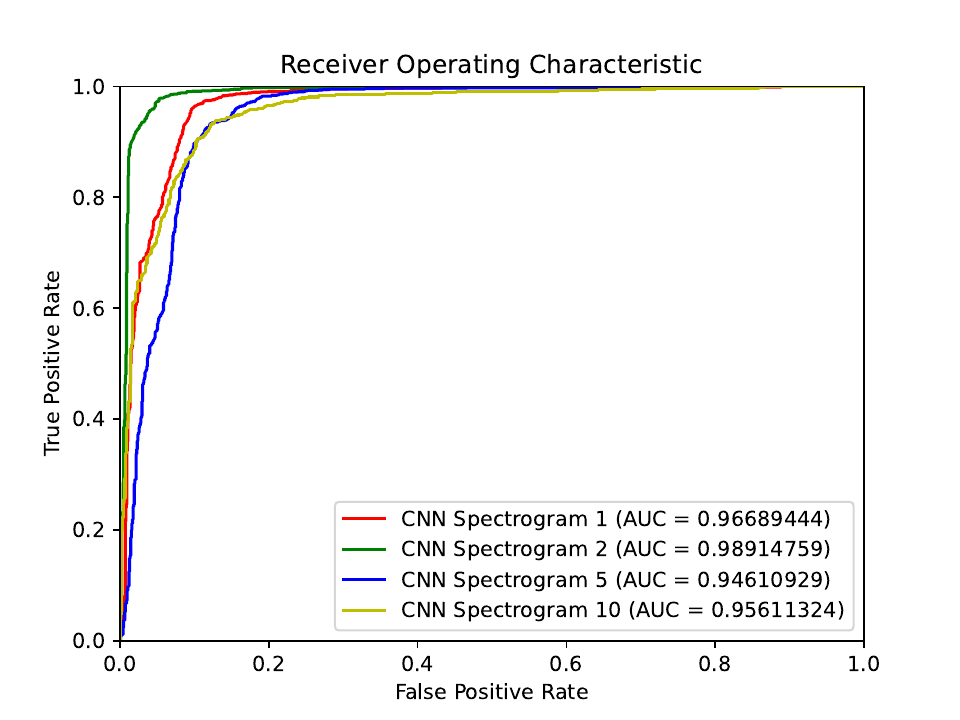}
        \caption{ROC and AUC CNN spectrogram}
        \label{fig:cnn_roc_spec}
    \end{subfigure}
    \caption{ROC and AUC for CNN models}
    \label{fig:cnn_roc_auc}
\end{figure}

Similarly, Figure \ref{fig:cnn_roc_time} presents the ROC curve and AUC for CNN models using the time-domain representation, Figure \ref{fig:cnn_roc_welch} for the Welch method, Figure \ref{fig:cnn_roc_multi} for the Multitaper method, and Figure \ref{fig:cnn_roc_spec} for the spectrogram with Multitaper.

\begin{figure}[htbp]
    \centering
    \begin{subfigure}{0.49\textwidth}
        \centering
        \includegraphics[width=\linewidth]{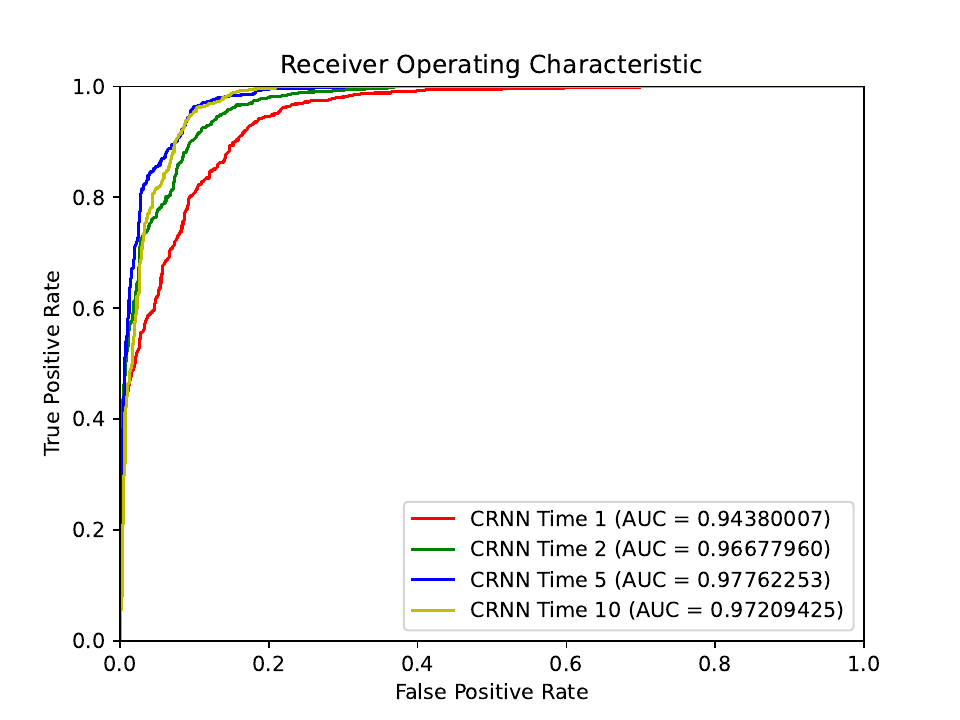}
        \caption{ROC and AUC CRNN time}
        \label{fig:crnn_roc_time}
    \end{subfigure}
    \hfill
    \begin{subfigure}{0.49\textwidth}
        \centering
        \includegraphics[width=\linewidth]{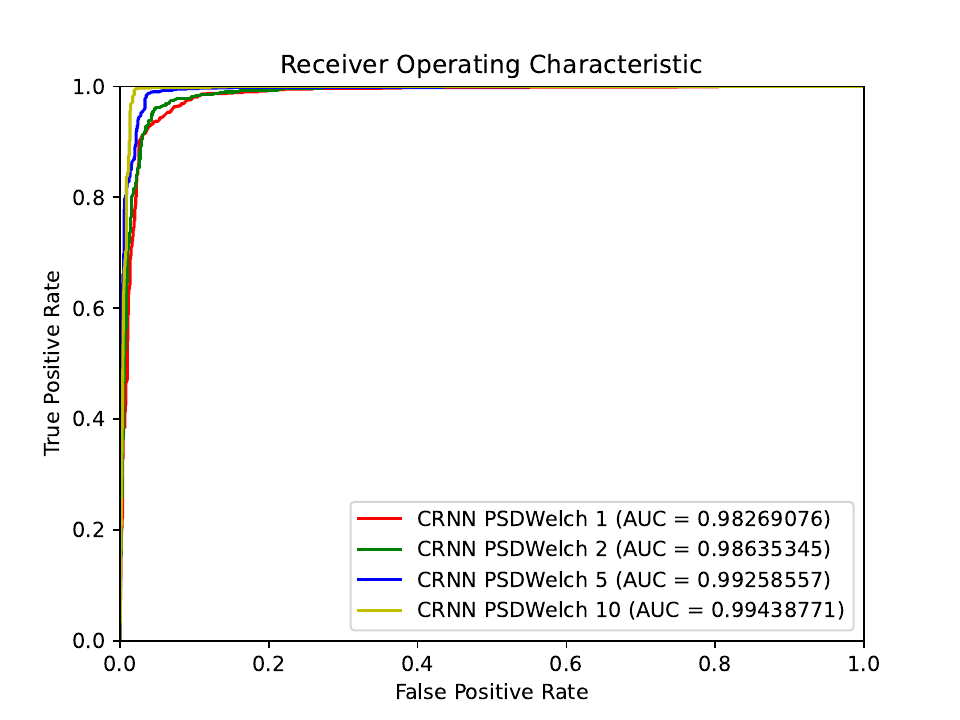}
        \caption{ROC and AUC CRNN Welch}
        \label{fig:crnn_roc_welch}
    \end{subfigure}
    \begin{subfigure}{0.49\textwidth}
        \centering
        \includegraphics[width=\linewidth]{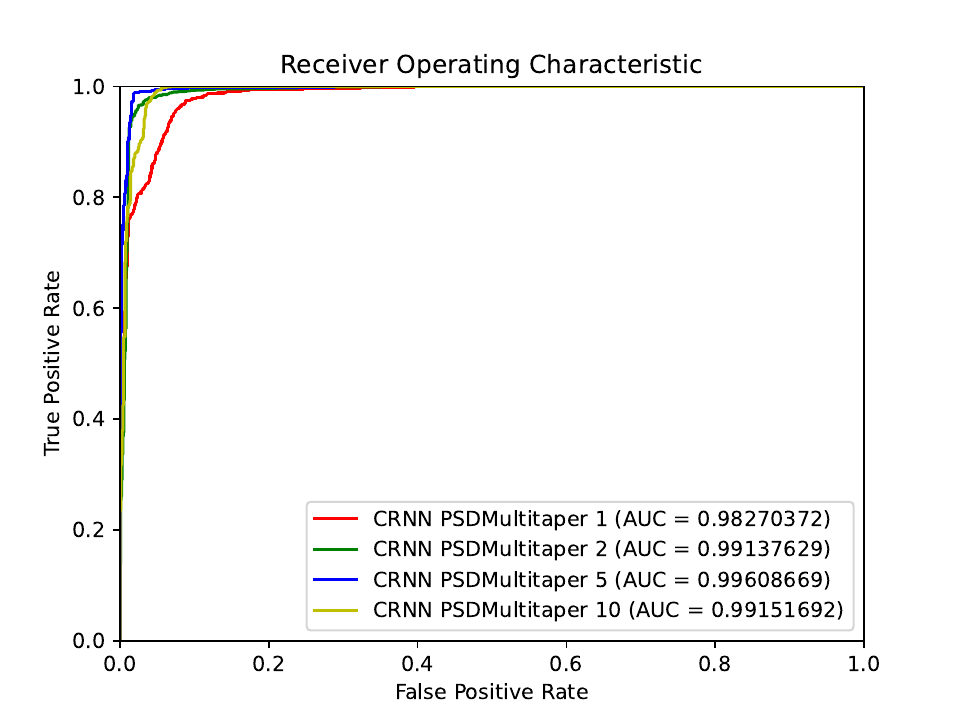}
        \caption{ROC and AUC CRNN Multitaper}
        \label{fig:crnn_roc_multi}
    \end{subfigure}
    \hfill
    \begin{subfigure}{0.49\textwidth}
        \centering
        \includegraphics[width=\linewidth]{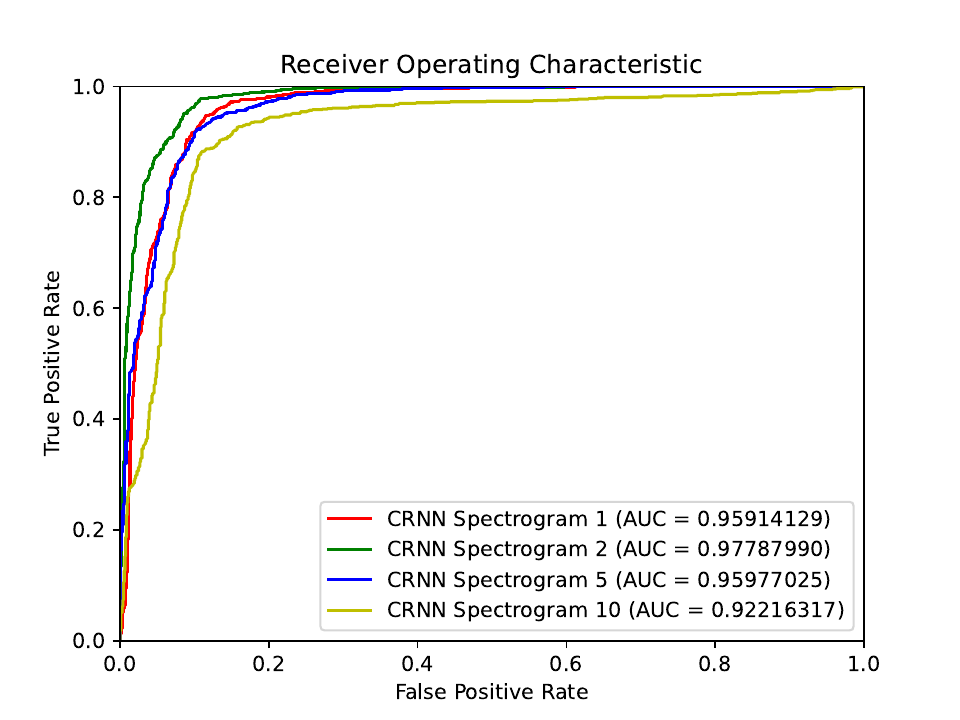}
        \caption{ROC and AUC CRNN spectrogram}
        \label{fig:crnn_roc_spec}
    \end{subfigure}
    \caption{ROC and AUC for CRNN models}
    \label{fig:crnn_roc_auc}
\end{figure}

For CRNN models, Figure \ref{fig:crnn_roc_time} shows the ROC curve and AUC for the time domain, Figure \ref{fig:crnn_roc_welch} for the Welch method, Figure \ref{fig:crnn_roc_multi} for the Multitaper method, and Figure \ref{fig:crnn_roc_spec} for the spectrogram with Multitaper.

Table \ref{tab:model_rank_by_auc} presents the ranking of models based on AUC, where higher AUC values indicate better classification performance.

\begin{table}[htbp]
\caption{Rank of models by AUC}
\centering
\begin{tabular}{|c|c|c|c|c|}
\hline
\textbf{Rank} & \textbf{Model} & \textbf{Signal Type} & \textbf{Window Size} & \textbf{AUC} \\ \hline
1  & CNN  & PSD Multitaper & 10s & 0.99871438 \\ \hline
2  & CNN  & PSD Multitaper & 5s  & 0.99793805 \\ \hline
3  & RNN  & PSD Multitaper & 10s & 0.99703315 \\ \hline
4  & CNN  & PSD Multitaper & 2s  & 0.99674661 \\ \hline
5  & CRNN & PSD Multitaper & 5s  & 0.99608669 \\ \hline
6  & RNN  & PSD Multitaper & 5s  & 0.99500515 \\ \hline
7  & CRNN & PSD Welch      & 10s & 0.99438771 \\ \hline
8  & RNN  & PSD Multitaper & 2s  & 0.99354539 \\ \hline
9  & CRNN & PSD Welch      & 5s  & 0.9924648  \\ \hline
10 & RNN  & PSD Welch      & 5s  & 0.99272361 \\ \hline
...& ...  & ...            & ... & ...        \\ \hline
40 & CRNN & Time           & 1s  & 0.94380007 \\ \hline
41 & RNN  & Spectrogram    & 5s  & 0.93729005 \\ \hline
42 & RNN  & Spectrogram    & 2s  & 0.93429124 \\ \hline
43 & RNN  & Time           & 2s  & 0.92392350 \\ \hline
44 & CRNN & Spectrogram    & 10s & 0.92216317 \\ \hline
45 & RNN  & Time           & 1s  & 0.91426225 \\ \hline
46 & RNN  & Spectrogram    & 10s & 0.90687733 \\ \hline
47 & RNN  & Time           & 10s & 0.86394842 \\ \hline
48 & RNN  & Time           & 5s  & 0.86178559 \\ \hline
\end{tabular}
\label{tab:model_rank_by_auc}
\end{table}

From the Table and Figures we observe that for all models (CNN, RNN, and CRNN), the AUC with the Multitaper method, reached almost perfect discrimination with AUC values above 0.99. This indicates an ideal false positive rate, signifying that the Multitaper method yields the best model performance in terms of distinguishing between seizure and non-seizure events. 

Following the Multitaper method, the Welch method (another frequency domain approach) also shows strong performance, with models utilizing Welch producing AUC values just slightly lower than Multitaper. This suggests that frequency-domain representations, particularly PSD Multitaper and Welch, consistently outperform time domain representations.

While still providing useful information, the spectrogram-based models perform somewhat less favorably than Multitaper and Welch. This indicates that although spectrogram analysis offers valuable time-frequency information, it may not capture the same level of discriminative power of frequency-domain methods.

Also, models that rely on time-domain signals show the lowest AUC values. This underscores the limited ability of raw time-domain signals to capture the relevant patterns and reinforce that feature extraction methods and preprocessing techniques in this type of data are needed.

\subsection{Statistical Analysis}

To identify the most effective model configurations, statistical analyses were performed based on model accuracy. These analyses ensure that differences between models are not only observable numerically but also statistically significant. In this analysis, we evaluate the differences between window groups, representation groups, and architecture groups separately.

\subsubsection{Kruskal-Wallis for window sizes}

The Kruskal-Wallis test was performed to compare the model groups by window size, for which, in each comparison, the same artificial neural architecture and signal representation are considered, resulting in 12 hypothesis tests. The results are summarized in Table \ref{tab:kruskal_results_windows}.

\begin{table}[htbp]
\centering
\caption{Kruskal-Wallis test results for different window sizes. }
\label{tab:kruskal_results_windows}
\begin{tabular}{|c|c|c|c|}
\hline
\textbf{Neural Network} & \textbf{Signal Type} & \textbf{p-value} \\ \hline
RNN  & Time           & 0.00228107 \\ \hline
RNN  & PSD Welch      & 0.00000014 \\ \hline
RNN  & PSD Multitaper & 0.00000006 \\ \hline
RNN  & Spectrogram    & 0.00005723 \\ \hline
CNN  & Time           & 0.00000119 \\ \hline
CNN  & PSD Welch      & 0.00000006 \\ \hline
CNN  & PSD Multitaper & 0.00000006 \\ \hline
CNN  & Spectrogram    & 0.00000145 \\ \hline
CRNN & Time           & 0.00000027 \\ \hline
CRNN & PSD Welch      & 0.00000006 \\ \hline
CRNN & PSD Multitaper & 0.00000006 \\ \hline
CRNN & Spectrogram    & 0.00000596 \\ \hline
\end{tabular}%
\end{table}

According to Table \ref{tab:kruskal_results_windows}, all comparisons resulted in statistical significant difference, whose p-values are lower than 0.05. Thus, the post-hoc analysis using Dunn's test was conducted to determine which specific window size comparisons led to significant differences in model performance. Statistically significant differences (p < 0.05) are highlighted in green. All tables are presented in ~\ref{apdx:statistical_ws}.

\begin{itemize}
\item \textbf{Table \ref{dunn_rnn_time}:} Significant differences in 10s vs. 2s and 2s vs. 5s;
\item \textbf{Table \ref{dunn_rnn_welch}:} Significant differences in 1s vs. 10s, 1s vs. 5s, and 10s vs. 2s;
\item \textbf{Table \ref{dunn_rnn_mult}:} Significant differences in 1s vs. 10s, 1s vs. 5s, and 10s vs. 2s;
\item \textbf{Table \ref{dunn_rnn_spec}:} Significant differences in 1s vs. 10s, 1s vs. 5s, 10s vs. 2s, and 2s vs. 5s;
\end{itemize}

The post-hoc analysis revealed significant differences in model performance across different window sizes. Notably, the largest discrepancies were observed in comparisons involving the shortest (1s and 2s) and longest (10s and 5s) windows, particularly in Table \ref{dunn_rnn_welch}, Table \ref{dunn_rnn_mult}, and Table \ref{dunn_rnn_spec}, suggesting that window size strongly influences model behavior.

\begin{itemize}
\item \textbf{Table \ref{dunn_cnn_time}:} Significant differences in 1s vs. 2s, 1s vs. 5s, 10s vs. 5s, and 2s vs. 5s;
\item \textbf{Table \ref{dunn_cnn_welch}:} Significant differences in 1s vs. 10s and 1s vs. 5s;
\item \textbf{Table \ref{dunn_cnn_mult}:} Significant differences in 1s vs. 10s and 1s vs. 5s;
\item \textbf{Table \ref{dunn_cnn_spec}:} Significant differences in 1s vs. 10s, 1s vs. 2s, and 2s vs. 5s;
\end{itemize}

The post-hoc analysis revealed that CNN models are affected by window size variations, with significant differences mostly involving the shortest (1s and 2s) and longest (10s and 5s) windows. The largest performance gaps appear in Table \ref{dunn_cnn_welch} and Table \ref{dunn_cnn_mult}, where 1s performs notably different from 10s and 5s.

\begin{itemize}
\item \textbf{Table \ref{dunn_crnn_time}:} Significant differences in 1s vs. 10s, 1s vs. 5s, and 10s vs. 2s;
\item \textbf{Table \ref{dunn_crnn_welch}:} Significant differences in 1s vs. 10s and 1s vs. 5s;
\item \textbf{Table \ref{dunn_crnn_mult}:} Significant differences in 1s vs. 10s and 1s vs. 5s;
\item \textbf{Table \ref{dunn_crnn_spec}:} Significant differences in 1s vs. 10s, 1s vs. 2s, and 2s vs. 5s;
\end{itemize}

The post-hoc analysis for CRNN models reveals a pattern similar to that observed in CNN models. The most pronounced effects occur in Table \ref{dunn_crnn_welch} and Table \ref{dunn_crnn_mult}, where the 1s window significantly differs from both 10s and 5s.

In conclusion, window size has a significant impact on model performance across RNN, CNN, and CRNN architectures and signal representations. Across all models, the shortest window sizes (1s and 2s) showed substantial differences when compared to the longest window sizes (5s and 10s), with these differences being most prominent in the frequency-domain representations (PSD Welch and PSD Multitaper). Reinforcing that window size plays a critical role in shaping model behavior.

\subsubsection{Kruskal-Wallis for signal representations}

Next, we performed the Kruskal-Wallis test to compare models by signal representation groups. The results of this test are summarized in Table \ref{tab:kruskal_results_representations}, which shows extremely significant differences in the accuracy of models and window sizes across different representations.

\begin{table}[htbp]
\centering
\caption{Kruskal-Wallis test results for different signal representations and window sizes}
\begin{tabular}{|c|c|c|c|}
\hline
\textbf{Neural Network} & \textbf{Window Size} & \textbf{p-value} \\ \hline
RNN   & 1s  & 0.00000020 \\ \hline
RNN   & 2s  & 0.00000025 \\ \hline
RNN   & 5s  & 0.00000006 \\ \hline
RNN   & 10s & 0.00000016 \\ \hline
CNN   & 1s  & 0.00000017 \\ \hline
CNN   & 2s  & 0.00000029 \\ \hline
CNN   & 5s  & 0.00000026 \\ \hline
CNN   & 10s & 0.00000007 \\ \hline
CRNN  & 1s  & 0.00000023 \\ \hline
CRNN  & 2s  & 0.00000011 \\ \hline
CRNN  & 5s  & 0.00000012 \\ \hline
CRNN  & 10s & 0.00000012 \\ \hline
\end{tabular}
\label{tab:kruskal_results_representations}
\end{table}

To assess pairwise differences, Dunn's post-hoc test was conducted for each combination. The test aimed to determine which specific representation led to significant differences. Statistically significant differences (p < 0.05) are highlighted in green. All tables are presented in ~\ref{apdx:statistical_rep}.

\begin{itemize}
\item \textbf{Table \ref{dunn_rnn_1s}:} Significant differences in Spectrogram vs. PSDMultitaper, Time vs. PSDMultitaper and Time vs. PSDWelch;
\item \textbf{Table \ref{dunn_rnn_2s}:} Significant differences in Spectrogram vs. PSDMultitaper, Time vs. PSDMultitaper and Time vs. PSDWelch;
\item \textbf{Table \ref{dunn_rnn_5s}:} Significant differences in Spectrogram vs. PSDMultitaper, Time vs. PSDMultitaper and Time vs. PSDWelch;
\item \textbf{Table \ref{dunn_rnn_10s}:} Significant differences in Spectrogram vs. PSDMultitaper, Time vs. PSDMultitaper and Time vs. PSDWelch;
\end{itemize}

For the RNN architecture, no significant differences were found between the frequency-domain representations (PSDMultitaper and PSDWelch). However, significant differences were observed between the time-domain representation (Time) and both frequency-domain representations (PSDMultitaper and PSDWelch). Similarly, the time-frequency representation (Spectrogram) showed significant differences when compared to the frequency-domain representation (PSDMultitaper).

\begin{itemize}
\item \textbf{Table \ref{dunn_cnn_1s}:} Significant differences in Spectrogram vs. PSDMultitaper, Time vs. PSDMultitaper and Time vs. PSDWelch;
\item \textbf{Table \ref{dunn_cnn_2s}:} Significant differences in PSDWelch vs. PSDMultitaper, Time vs. PSDMultitaper and Spectrogram vs. Time;
\item \textbf{Table \ref{dunn_cnn_5s}:} Significant differences in Spectrogram vs. PSDMultitaper, Spectrogram vs. PSDWelch and Time vs. PSDMultitaper;
\item \textbf{Table \ref{dunn_cnn_10s}:} Significant differences in Spectrogram vs. PSDMultitaper, Time vs. PSDMultitaper and Time vs. PSDWelch;
\end{itemize}

For the CNN models, consistent significant differences were observed between the frequency-domain representations (PSDMultitaper and PSDWelch) when compared to the time-domain (Time) and time-frequency (Spectrogram) representations. However, no significant differences were found between the frequency-domain representations themselves, except for the 2-second window.

\begin{itemize}
\item \textbf{Table \ref{dunn_crnn_1s}:} Significant differences in Spectrogram vs. PSDMultitaper, Time vs. PSDMultitaper and Time vs. PSDWelch;
\item \textbf{Table \ref{dunn_crnn_2s}:} Significant differences in PSDWelch vs. PSDMultitaper, Time vs. PSDMultitaper and Spectrogram vs. Time;
\item \textbf{Table \ref{dunn_crnn_5s}:} Significant differences in Spectrogram vs. PSDMultitaper, Spectrogram vs. PSDWelch and Time vs. PSDMultitaper;
\item \textbf{Table \ref{dunn_crnn_10s}:} Significant differences in Spectrogram vs. PSDMultitaper, Spectrogram vs. PSDWelch and Time vs. PSDMultitaper;
\end{itemize}

Similar to the CNN and RNN models, the CRNN models showed significant differences between the time-domain (Time) and frequency-domain (PSDMultitaper and PSDWelch) representations. Furthermore, the time-frequency representation (Spectrogram) showed significant differences when compared to the frequency-domain representation.

In summary, the post-hoc comparisons across the RNN, CNN, and CRNN models reveal consistent trends in the significant differences among time-domain, frequency-domain, and time-frequency domain representations. While the frequency-domain representations (PSD Multitaper and PSD Welch) showed few significant differences between themselves, they consistently differed from both the time-domain (Time) and time-frequency (Spectrogram) representations. 

\subsubsection{Friedman for model architecture}

The Friedman test was conducted to evaluate whether there were significant differences in the model architecture group. The results, presented in Table \ref{tab:friedman}, include the H-values and p-values for each signal type and window size. A p-value below 0.05 indicates a statistically significant difference. Values with no statistically significant difference are highlighted in red.

\begin{table}[htbp]
\centering
\caption{Friedman test results for each signal representation and window size.}
\label{tab:friedman}
\begin{tabular}{|c|c|c|c|}
\hline
\textbf{Signal Type} & \textbf{Window Size} & \textbf{p-value} \\ \hline
Time            & 1s  & 0.00050045                \\ \hline
Time            & 2s  & 0.00004540                \\ \hline
Time            & 5s  & 0.00037074                \\ \hline
Time            & 10s & 0.00004540                \\ \hline
PSD Welch       & 1s  & \cellcolor{red}0.61435675 \\ \hline
PSD Welch       & 2s  & 0.00004540                \\ \hline
PSD Welch       & 5s  & 0.00022487                \\ \hline
PSD Welch       & 10s & 0.03826736                \\ \hline
PSD Multitaper  & 1s  & 0.03826736                \\ \hline
PSD Multitaper  & 2s  & 0.00037074                \\ \hline
PSD Multitaper  & 5s  & 0.00055308                \\ \hline
PSD Multitaper  & 10s & 0.00123091                \\ \hline
Spectrogram     & 1s  & \cellcolor{red}0.49658530 \\ \hline
Spectrogram     & 2s  & 0.00011167                \\ \hline
Spectrogram     & 5s  & 0.00037074                \\ \hline
Spectrogram     & 10s & 0.00050045                \\ \hline
\end{tabular}
\end{table}

To identify pairwise differences, Nemenyi post-hoc test was conducted for each combination. The test aimed to determine which specific representation led to significant differences. Statistically significant differences (p < 0.05) are highlighted in green. Combinations where the Friedman test did not show statistically significant differences are not included in the post-hoc analysis. All tables are presented in ~\ref{apdx:statistical_arc}.

\begin{itemize}
\item \textbf{Table \ref{friedman_time_1s}:} Significant differences in CNN vs. RNN and CRNN vs. RNN;
\item \textbf{Table \ref{friedman_time_2s}:} Significant differences in CNN vs. CRNN, CNN vs. RNN and CRNN vs. RNN;
\item \textbf{Table \ref{friedman_time_5s}:} Significant differences in CNN vs. CRNN, CNN vs. RNN and CRNN vs. RNN;
\item \textbf{Table \ref{friedman_time_10s}:} Significant differences in CNN vs. CRNN, CNN vs. RNN and CRNN vs. RNN;
\end{itemize}

\begin{itemize}   
\item \textbf{Table \ref{friedman_psdwelch_2s}:} Significant differences in CNN vs. CRNN, CNN vs. RNN and CRNN vs. RNN;
\item \textbf{Table \ref{friedman_psdwelch_5s}:} Significant differences in CNN vs. CRNN, CNN vs. RNN and CRNN vs. RNN;
\item \textbf{Table \ref{friedman_psdwelch_10s}:} Significant differences in CNN vs. CRNN and CRNN vs. RNN;
\end{itemize}

\begin{itemize}   
\item \textbf{Table \ref{friedman_psdmultitaper_1s}:} Significant differences in CNN vs. RNN;
\item \textbf{Table \ref{friedman_psdmultitaper_2s}:} Significant differences in CNN vs. CRNN, CNN vs. RNN and CRNN vs. RNN;
\item \textbf{Table \ref{friedman_psdmultitaper_5s}:} Significant differences CNN vs. CRNN and CNN vs. RNN;
\item \textbf{Table \ref{friedman_psdmultitaper_10s}:} Significant differences CNN vs. CRNN and CNN vs. RNN;
\end{itemize}

\begin{itemize}   
\item \textbf{Table \ref{friedman_spectrogram_2s}:} Significant differences in CNN vs. RNN and CRNN vs. RNN;
\item \textbf{Table \ref{friedman_spectrogram_5s}:} Significant differences in CNN vs. RNN and CRNN vs. RNN;
\item \textbf{Table \ref{friedman_spectrogram_10s}:} Significant differences in CNN vs. RNN and CRNN vs. RNN;
\end{itemize}

The Friedman and Nemenyi tests indicate that model architecture significantly influences performance, with CNN models consistently differing from RNN and CRNN.

\subsection{Wrong Prediction Analysis}

In this analysis, we aim to understand the factors contributing to erroneous detections in certain types of signals. Specifically, we investigate whether these misclassifications originate from preprocessing steps, incorrect ground truth labels in the dataset, or inherent limitations in the model’s ability to classify all signal variations accurately.

To conduct this analysis, we first identified a file where most models exhibited incorrect predictions. The selected record from the dataset is chb04, file 28, which contains the following seizure-related annotations:

\begin{itemize}
    \item \textbf{Number of seizures:} 2
    \item \textbf{Seizure 1:} Start: 1679s, End: 1781s, Duration: 102s
    \item \textbf{Seizure 2:} Start: 3782s, End: 3898s, Duration: 116s
\end{itemize}

\begin{figure}
    \centering
    \includegraphics[width=\linewidth]{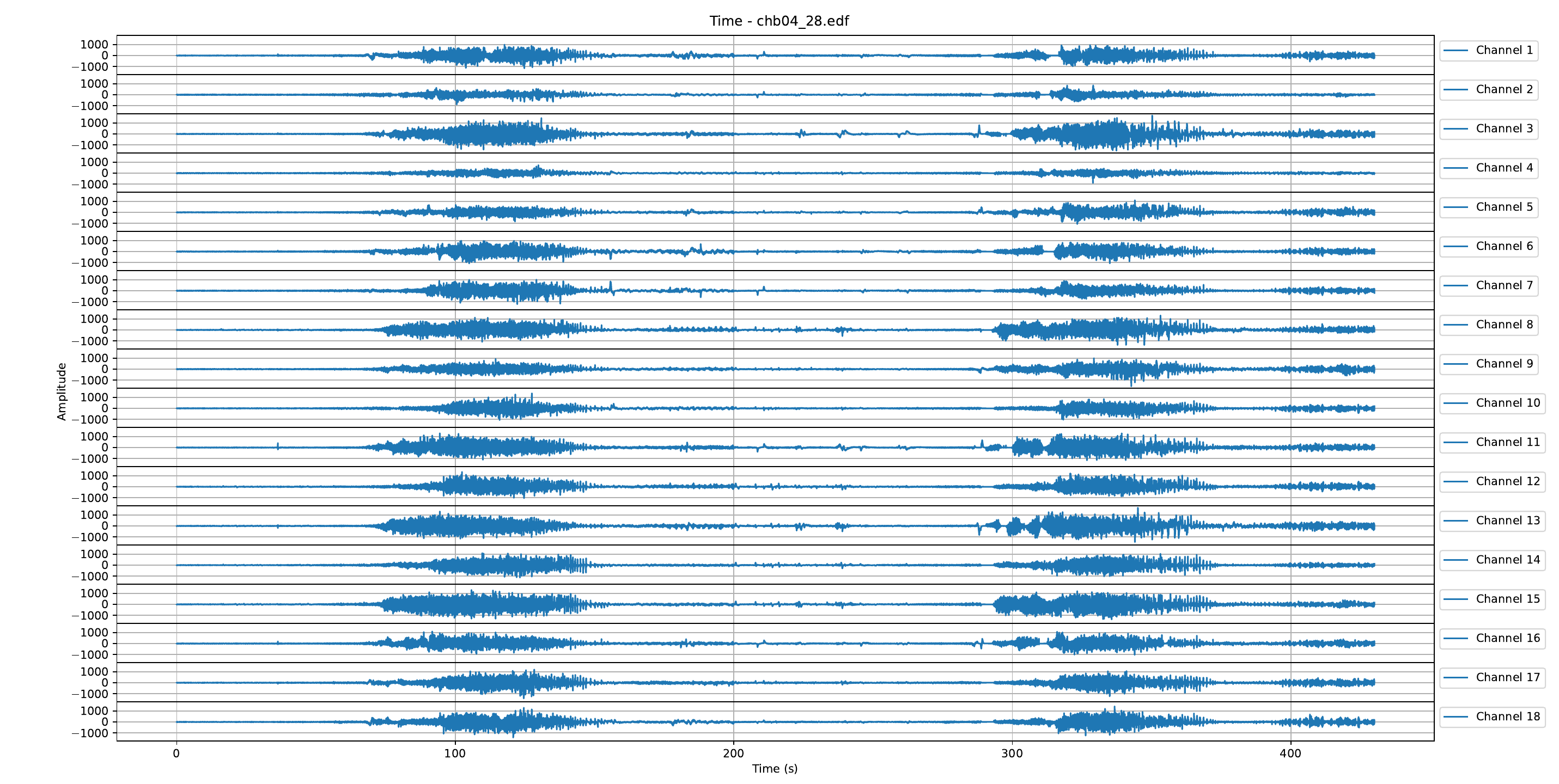}
    \caption{Time signal generated from file chb04-28.}
    \label{fig:chb04_28}
\end{figure}

For this evaluation, we selected the models with the highest performance metrics, namely the CNN, RNN, and CRNN trained with a 10-second window and the PSD representation generated by the Multitaper method. The preprocessing of this file involved segmenting it into 10-second, non-overlapping windows, allowing us to examine specific windows and corresponding model predictions. This process yielded 432 seconds of EEG activity, divided into 43 segmented windows. Figure \ref{fig:chb04_28} shows the time signal from the selected file.

Each evaluated model, trained using 10-fold cross-validation, was used to generate predictions for all 43 windows. Table \ref{tab:wrong_predictions} presents the misclassified instances, including the indices of erroneously predicted windows, their start and end times, and the number of misclassifications per architecture. Green-highlighted cells indicate non-seizure segments, yellow-highlighted cells indicate windows containing seizure information, and red-highlighted cells represent edge windows, which include both seizure and non-seizure data, potentially affecting classification performance as they are considered part of the seizure segment.

\begin{table}[h]
\caption{Number of occurrences of each misclassified index per model type}
\centering
\begin{tabular}{|c|c|c|c|c|c|}
\hline
\textbf{Index} & \textbf{Start and End} & \textbf{Total} & \textbf{RNN} & \textbf{CNN} & \textbf{CRNN} \\
\hline
4  & \cellcolor{green}1628 - 1638  & 6  & 6 & 0 & 0 \\
5  & \cellcolor{green}1638 - 1648  & 2  & 0 & 0 & 2 \\
14 & \cellcolor{yellow}1768 - 1778 & 1  & 0 & 0 & 1 \\
15 & \cellcolor{red}1778 - 1788    & 13 & 0 & 4 & 9 \\
20 & \cellcolor{green}3724 - 3734  & 1  & 0 & 0 & 1 \\
25 & \cellcolor{red}3784 - 3794    & 17 & 3 & 7 & 7 \\
26 & \cellcolor{yellow}3794 - 3804 & 1  & 0 & 0 & 1  \\
27 & \cellcolor{yellow}3804 - 3814 & 2  & 0 & 1 & 1 \\
28 & \cellcolor{yellow}3814 - 3824 & 3  & 3 & 0 & 0 \\
37 & \cellcolor{green}3904 - 3914  & 1  & 0 & 0 & 1  \\
38 & \cellcolor{green}3914 - 3924  & 2  & 2 & 0 & 0 \\
\hline
\end{tabular}    
\label{tab:wrong_predictions}
\end{table}

From Table \ref{tab:wrong_predictions}, it is evident that edge windows represent the most significant challenge for all models. Specifically, Index 25 (3784-3794s) stands out, where these windows lead to the highest number of misclassifications due to the presence of both epileptic and non-epileptic data within the same segment. The CRNN model appears to face the most difficulty with edge windows, particularly in \textit{Index 15} and \textit{Index 25}, where it recorded the highest number of misclassifications.

For seizure windows, the RNN and CRNN models show a similar number of misclassifications (3 occurrences), while the CNN only made one misclassification. For non-seizure windows, the RNN model performed the worst, with 8 misclassifications, followed by CRNN with 4, and CNN did not misclassify any non-seizure windows.

\begin{figure}[htbp]
    \centering
    \begin{subfigure}{\textwidth}
        \centering
        \includegraphics[width=\linewidth]{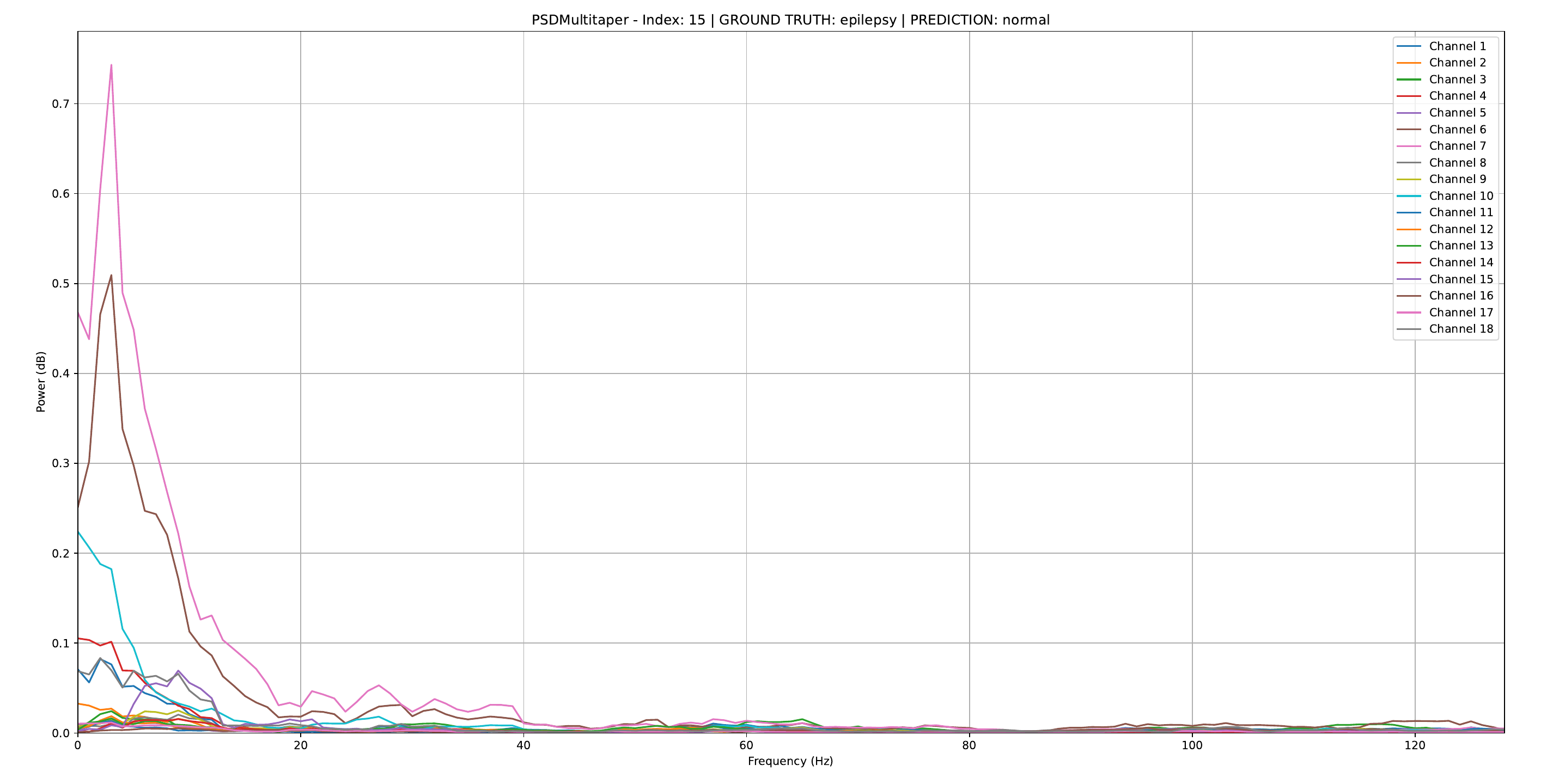}
        \caption{PSD for index 15}
        \label{fig:index15_psd}
    \end{subfigure}
    \begin{subfigure}{\textwidth}
        \centering
        \includegraphics[width=\linewidth]{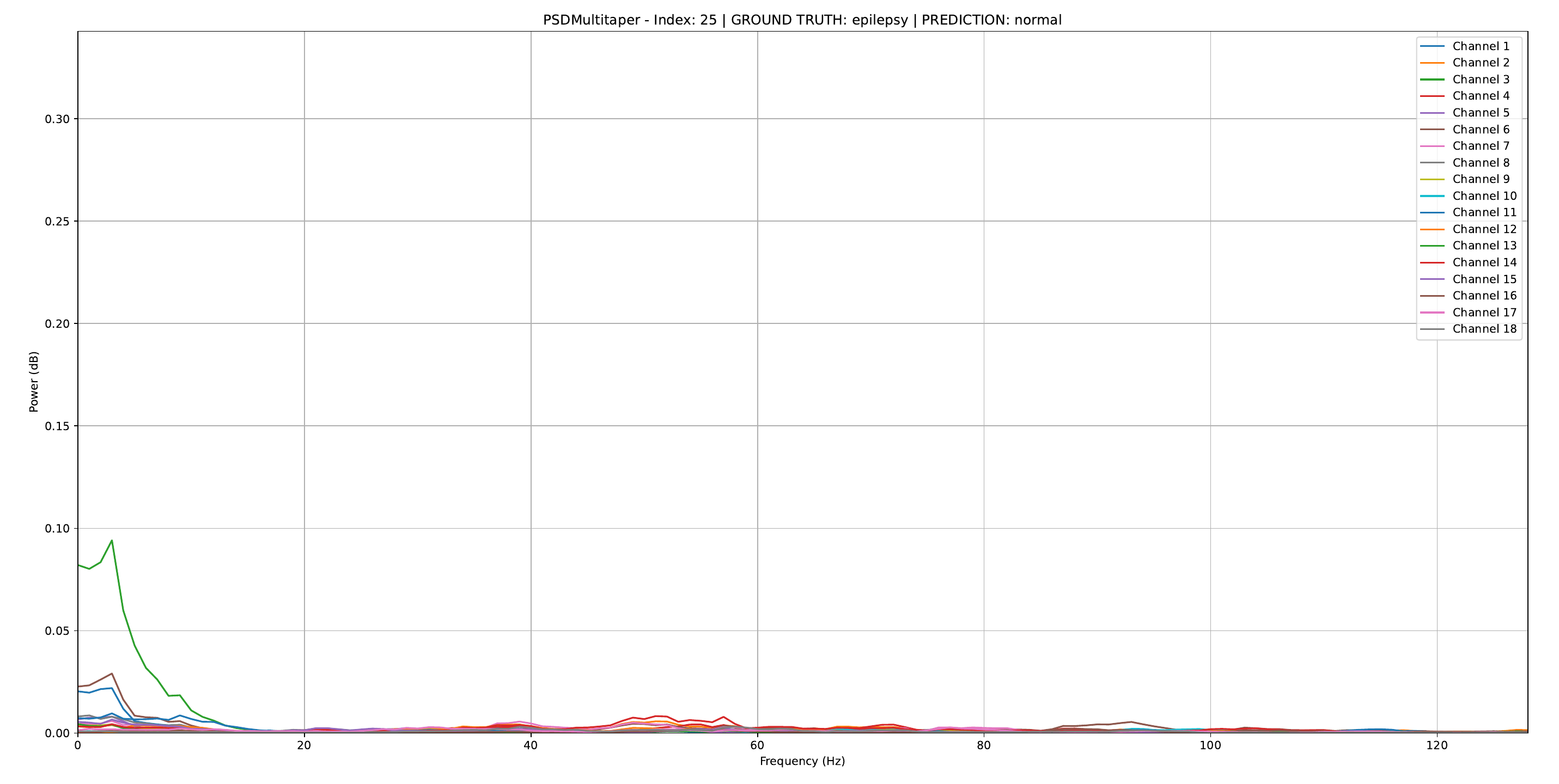}
        \caption{PSD for index 25}
        \label{fig:index25_psd}
    \end{subfigure}
    \caption{Edge windows}
    \label{fig:edge_windows}
\end{figure}

To better visualize these data, Figure \ref{fig:edge_windows} illustrates the edge windows (\textit{Index 15} and \textit{Index 25}). Sub-figures \ref{fig:index15_psd} and \ref{fig:index25_psd} present their respective PSD graphs.

\begin{figure}[htbp]
    \centering
    \begin{subfigure}{\textwidth}
        \centering
        \includegraphics[width=\linewidth]{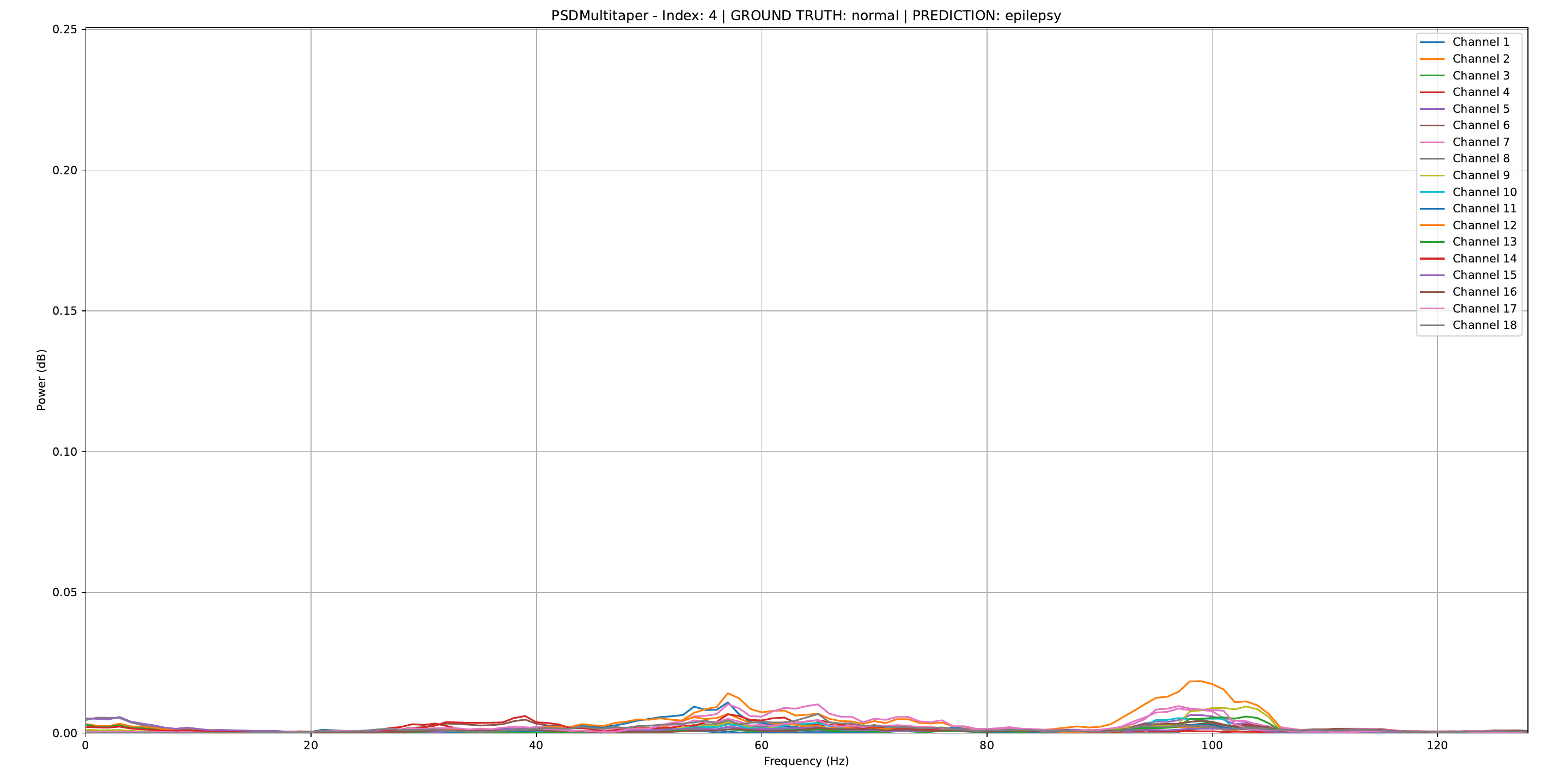}
        \caption{PSD for index 04}
        \label{fig:index04_psd}
    \end{subfigure}
    \caption{Non-seizure windows}
    \label{fig:non_seizure_windows}
\end{figure}

\begin{figure}[htbp]
    \centering
    \begin{subfigure}{\textwidth}
        \centering
        \includegraphics[width=\linewidth]{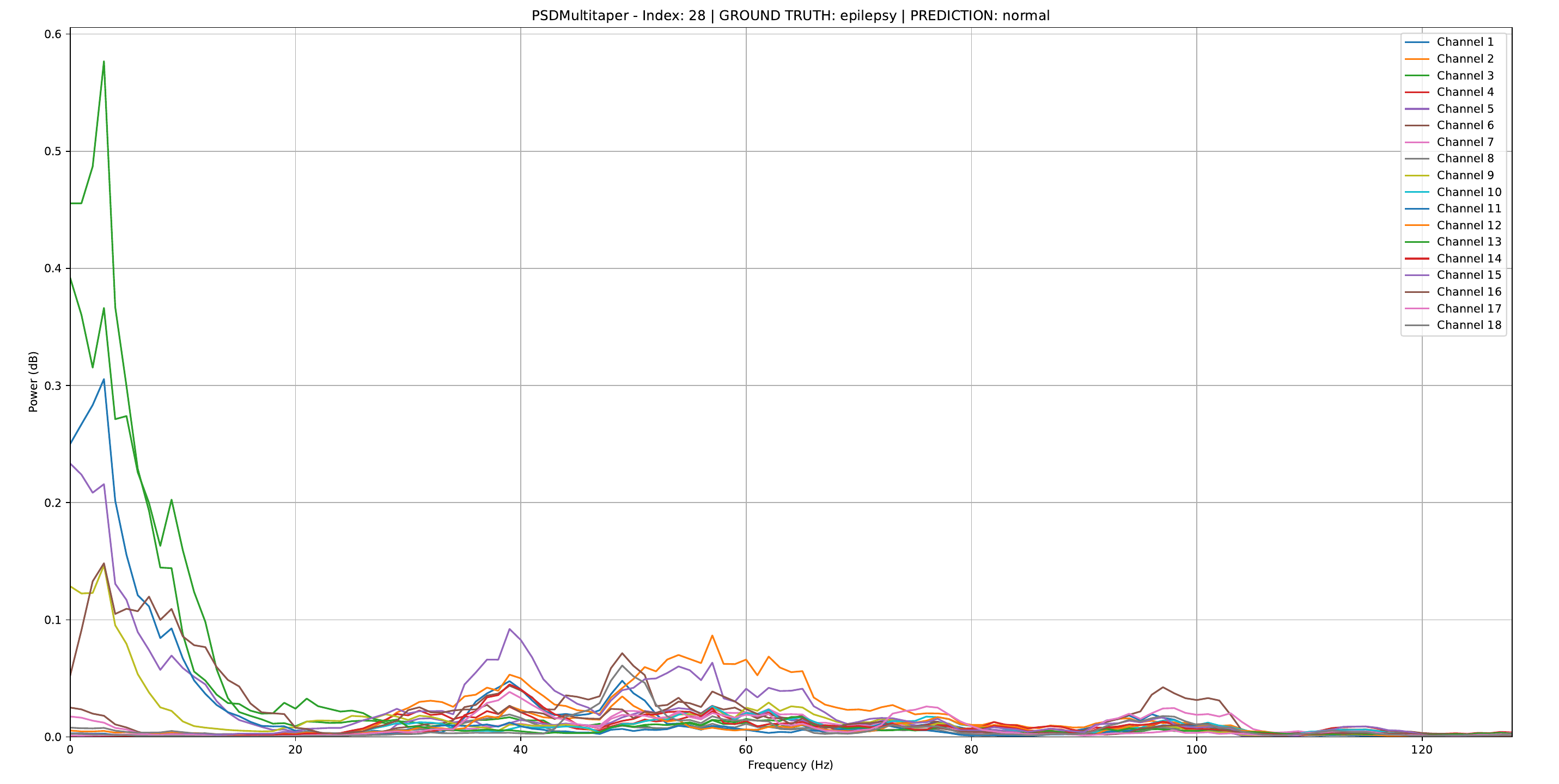}
        \caption{PSD for index 28}
        \label{fig:index28_psd}
    \end{subfigure}
    \caption{Seizure windows}
    \label{fig:seizure_windows}
\end{figure}

If we closely examine both \textit{Index 25} and \textit{Index 15}, we observe distinct patterns in their PSD data. \textit{Index 15} exhibits characteristics more similar to a seizure window, as seen in \textit{Index 28} in Figure \ref{fig:seizure_windows}, whereas \textit{Index 25} resembles a non-seizure window, similar to \textit{Index 04} in Figure \ref{fig:non_seizure_windows}. This reinforces the idea that edge cases exhibit unique characteristics and may require a distinct treatment from both seizure and non-seizure segments to improve classification accuracy.

In general the CNN model achieved the best performance on both non-seizure and seizure windows, exhibiting fewer misclassifications overall. Although it still struggles with edge windows, its performance is comparatively superior to the other models.

Although this evaluation focuses on the PSD Multitaper representation with a 10-second window, edge-case misclassifications can occur across all representations used in this work.

To address the challenge of misclassifications in edge windows, one potential solution is to employ an ensemble model that integrates predictions from multiple models trained on varying segment sizes. This approach could enhance classification robustness by capturing both short and long term dependencies within the EEG signals. Additionally, weighting the contributions of each model based on their performance in different window types (seizure, non-seizure, and edge) could further refine classification accuracy.

Another alternative is to introduce a separate class specifically for edge windows. By explicitly labeling and training models to recognize this distinct category, the classification process could better differentiate between seizure and non-seizure patterns, potentially reducing misclassification rates.

\section{Conclusion}\label{section:CONCLUSION}

This study investigates the impact of various EEG data representations (time, frequency, and time-frequency domains) on the performance of deep learning models for epileptic seizure detection. Using the PhysioNet dataset, which consists of real EEG data, we trained and evaluated models across several representations and window sizes.

Our findings show that frequency-domain representations, particularly the PSD generated by the Multitaper and Welch methods, significantly outperform other representations across all metrics, window sizes, and models. Since both methods showed no significant differences in the statistical analyses, they can be considered interchangeable for epileptic seizure detection.

Window size was found to significantly influence model performance, with longer windows (5s and 10s) yielding improved results, especially in frequency-domain representations, as expected.

Architectural comparisons demonstrate the strong performance of CNN models, which consistently outperformed or matched more complex architectures like RNNs and CRNNs across various representations and window sizes. CNNs were particularly advantageous due to their computational efficiency, requiring less GPU memory compared to RNN and CRNN models, which are more demanding due to their temporal dependencies (e.g., LSTM layers). However, edge-case misclassifications remained a challenge for all models.

\subsection{Limitations}

The main limitations of this work are the following:

\begin{itemize}
    \item Only one dataset was used, limiting the generalizability of the results to other EEG datasets or seizure types.
    \item The proposed method was applied only to EEG signals related to epilepsy.
\end{itemize}

\subsection{Contributions}

This work conceived the following contributions:

\begin{itemize}
    \item A systematic comparison of EEG data representations (time, frequency, and time-frequency domains) for deep learning-based epileptic seizure detection.
    \item A demonstration of models architecture performance under different conditions for epileptic seizure detection.
    \item Insights into data representations and window sizes for maximizing seizure detection.
    \item A comparison of ROC and AUC curves to assess which model best minimizes false positive detections.
    \item A statistical comparison between models, data representations and window sizes to determine their impact on seizure detection.
\end{itemize}

\subsection{Future Works}

Future works include:

\begin{itemize} 
    \item Explore the application of the proposed method to other neurological conditions, such as Alzheimer's or dyslexia, to assess the generalizability of the approach.
    \item Investigating the performance of models with other deep learning architectures, such as Transformer-based models. 
    \item Expanding the dataset to include more diverse seizure types and patients, improving the robustness of the model.
    \item Employing techniques such as ensemble learning or multi-class classification to handle edge cases.
    \item Optimizing the computational efficiency of the models by investigating techniques such as pruning, quantization, or transfer learning.
\end{itemize}

\appendix

\section{Tables from pairwise comparison for window sizes}
\label{apdx:statistical_ws}

\begin{table}[htbp]
\centering
\caption{Dunn test results for RNN - Time}
\label{dunn_rnn_time}
\begin{tabular}{|c|c|c|c|c|}
\hline
\multicolumn{5}{|c|}{\textbf{Dunn test for RNN - Time}} \\
\hline
    & 1s & 10s & 2s & 5s \\ \hline
1s  & X & 0.381279 & 0.876219 & 0.576584 \\ \hline
10s & 0.381279 & X & \cellcolor{green}0.005617 & 1.000000 \\ \hline
2s  & 0.876219 & \cellcolor{green}0.005617 & X & \cellcolor{green}0.010934 \\ \hline
5s  & 0.576584 & 1.000000 & \cellcolor{green}0.010934 & X \\ \hline
\end{tabular}
\end{table}

\begin{table}[htbp]
\centering
\caption{Dunn test results for RNN - PSD Welch}
\label{dunn_rnn_welch}
\begin{tabular}{|c|c|c|c|c|}
\hline
\multicolumn{5}{|c|}{\textbf{Dunn test for RNN - PSD Welch}} \\
\hline
    & 1s & 10s & 2s & 5s \\ \hline
1s  & X & \cellcolor{green}0.000000 & 1.000000 & \cellcolor{green}0.002406 \\ \hline
10s & \cellcolor{green}0.000000 & X & \cellcolor{green}0.000234 & 0.334351 \\ \hline
2s  & 1.000000 & \cellcolor{green}0.000234 & X & 0.166775 \\ \hline
5s  & \cellcolor{green}0.002406 & 0.334351 & 0.166775 & X  \\ \hline
\end{tabular}
\end{table}

\begin{table}[htbp]
\centering
\caption{Dunn test results for RNN - PSD Multitaper}
\label{dunn_rnn_mult}
\begin{tabular}{|c|c|c|c|c|}
\hline
\multicolumn{5}{|c|}{\textbf{Dunn test for RNN - PSD Multitaper}} \\
\hline
    & 1s & 10s & 2s & 5s \\ \hline
1s  & X & \cellcolor{green}0.000000 & 0.334558 & \cellcolor{green}0.000782 \\ \hline
10s & \cellcolor{green}0.000000 & X & \cellcolor{green}0.000782 & 0.334558 \\ \hline
2s  & 0.334558 & \cellcolor{green}0.000782 & X & 0.334558 \\ \hline
5s  & \cellcolor{green}0.000782 & 0.334558 & 0.334558 & X  \\ \hline
\end{tabular}
\end{table}

\begin{table}[htbp]
\centering
\caption{Dunn test results for RNN - Spectrogram}
\label{dunn_rnn_spec}
\begin{tabular}{|c|c|c|c|c|}
\hline
\multicolumn{5}{|c|}{\textbf{Dunn test for RNN - Spectrogram}} \\
\hline
    & 1s & 10s & 2s & 5s \\ \hline
1s  & X & \cellcolor{green}0.002084 & 1.000000 & \cellcolor{green}0.029543 \\ \hline
10s & \cellcolor{green}0.002084 & X & \cellcolor{green}0.000986 & 1.000000 \\ \hline
2s  & 1.000000 & \cellcolor{green}0.000986 & X & \cellcolor{green}0.016026 \\ \hline
5s  & \cellcolor{green}0.029543 & 1.000000 & \cellcolor{green}0.016026 & X \\ \hline
\end{tabular}
\end{table}

\begin{table}[htbp]
\centering
\caption{Dunn test for CNN - Time}
\label{dunn_cnn_time}
\begin{tabular}{|c|c|c|c|c|}
\hline
\multicolumn{5}{|c|}{\textbf{Dunn test for CNN - Time}} \\
\hline
    & 1s & 10s & 2s & 5s \\ \hline
1s  & X & 0.202107 & \cellcolor{green}0.035190 & \cellcolor{green}0.000000 \\ \hline
10s & 0.202107 & X & 1.000000 & \cellcolor{green}0.005225 \\ \hline
2s  & \cellcolor{green}0.035190 & 1.000000 & X & \cellcolor{green}0.041875 \\ \hline
5s  & \cellcolor{green}0.000000 & \cellcolor{green}0.005225 & \cellcolor{green}0.041875 & X \\ \hline
\end{tabular}
\end{table}

\begin{table}[htbp]
\centering
\caption{Dunn test for CNN - PSD Welch}
\label{dunn_cnn_welch}
\begin{tabular}{|c|c|c|c|c|}
\hline
\multicolumn{5}{|c|}{\textbf{Dunn test for CNN - PSD Welch}} \\
\hline
    & 1s & 10s & 2s & 5s \\ \hline
1s  & X & \cellcolor{green}0.000000 & 0.334351 & \cellcolor{green}0.000780 \\ \hline
10s & \cellcolor{green}0.000000 & X & \cellcolor{green}0.000780 & 0.334351 \\ \hline
2s  & 0.334351 & \cellcolor{green}0.000780 & X & 0.334351 \\ \hline
5s  & \cellcolor{green}0.000780 & 0.334351 & 0.334351 & X \\ \hline
\end{tabular}
\end{table}

\begin{table}[htbp]
\centering
\caption{Dunn test for CNN - PSD Multitaper}
\label{dunn_cnn_mult}
\begin{tabular}{|c|c|c|c|c|}
\hline
\multicolumn{5}{|c|}{\textbf{Dunn test for CNN - PSD Multitaper}} \\
\hline
    & 1s & 10s & 2s & 5s \\ \hline
1s  & X & \cellcolor{green}0.000000 & 0.334489 & \cellcolor{green}0.000781 \\ \hline
10s & \cellcolor{green}0.000000 & X & \cellcolor{green}0.000781 & 0.334489 \\ \hline
2s  & 0.334489 & \cellcolor{green}0.000781 & X & 0.334489 \\ \hline
5s  & \cellcolor{green}0.000781 & 0.334489 & 0.334489 & X \\ \hline
\end{tabular}
\end{table}

\begin{table}[htbp]
\centering
\caption{Dunn test for CNN - Spectrogram}
\label{dunn_cnn_spec}
\begin{tabular}{|c|c|c|c|c|}
\hline
\multicolumn{5}{|c|}{\textbf{Dunn Test for CNN - Spectrogram}} \\
\hline
    & 1s & 10s & 2s & 5s \\ \hline
1s  & X & \cellcolor{green}0.017024 & \cellcolor{green}0.000000 & 0.151080 \\ \hline
10s & \cellcolor{green}0.017024 & X & 0.090602 & 1.000000 \\ \hline
2s  & \cellcolor{green}0.000000 & 0.090602 & X & \cellcolor{green}0.008954 \\ \hline
5s  & 0.151080 & 1.000000 & \cellcolor{green}0.008954 & X \\ \hline
\end{tabular}
\end{table}

\begin{table}[hhbp]
\centering
\caption{Dunn test results for CRNN - Time}
\label{dunn_crnn_time}
\begin{tabular}{|c|c|c|c|c|}
\hline
\multicolumn{5}{|c|}{\textbf{Dunn test for CRNN - Time}} \\
\hline
    & 1s & 10s & 2s & 5s \\ \hline
1s  & X & \cellcolor{green}0.000000 & 0.255555 & \cellcolor{green}0.000277 \\ \hline
10s & \cellcolor{green}0.000000 & X & \cellcolor{green}0.004891 & 1.000000 \\ \hline
2s  & 0.255555 & \cellcolor{green}0.004891 & X & 0.244059 \\ \hline
5s  & \cellcolor{green}0.000277 & 1.000000 & 0.244059 & X \\ \hline
\end{tabular}
\end{table}

\begin{table}[htbp]
\centering
\caption{Dunn test results for CRNN - PSD Welch}
\label{dunn_crnn_welch}
\begin{tabular}{|c|c|c|c|c|}
\hline
\multicolumn{5}{|c|}{\textbf{Dunn test for CRNN - PSD Welch}} \\
\hline
    & 1s & 10s & 2s & 5s \\ \hline
1s  & X  & \cellcolor{green}0.000000 & 0.334489 & \cellcolor{green}0.000781 \\ \hline
10s & \cellcolor{green}0.000000 & X  & \cellcolor{green}0.000781 & 0.334489 \\ \hline
2s  & 0.334489 & \cellcolor{green}0.000781 & X  & 0.334489 \\ \hline
5s  & \cellcolor{green}0.000781 & 0.334489 & 0.334489 & X \\ \hline
\end{tabular}
\end{table}

\begin{table}[htbp]
\centering
\caption{Dunn test results for CRNN - PSD Multitaper}
\label{dunn_crnn_mult}
\begin{tabular}{|c|c|c|c|c|}
\hline
\multicolumn{5}{|c|}{\textbf{Dunn test for CRNN - PSD Multitaper}} \\
\hline
    & 1s & 10s & 2s & 5s \\ \hline
1s  & X & \cellcolor{green}0.000000 & 0.334351 & \cellcolor{green}0.000780 \\ \hline
10s & \cellcolor{green}0.000000 & X & \cellcolor{green}0.000780 & 0.334351 \\ \hline
2s  & 0.334351 & \cellcolor{green}0.000780 & X & 0.334351 \\ \hline
5s  & \cellcolor{green}0.000780 & 0.334351 & 0.334351 & X \\ \hline
\end{tabular}
\end{table}

\begin{table}[htbp]
\centering
\caption{Dunn test results for CRNN - Spectrogram}
\label{dunn_crnn_spec}
\begin{tabular}{|c|c|c|c|c|}
\hline
\multicolumn{5}{|c|}{\textbf{Dunn test for CRNN - Spectrogram}} \\
\hline
    & 1s & 10s & 2s & 5s \\ \hline
1s  & X  & \cellcolor{green}0.008940 & \cellcolor{green}0.000002 & 0.136640 \\ \hline
10s & \cellcolor{green}0.008940 & X  & 0.319592 & 1.000000 \\ \hline
2s  & \cellcolor{green}0.000002 & 0.319592 & X  & \cellcolor{green}0.027740 \\ \hline
5s  & 0.136640 & 1.000000 & \cellcolor{green}0.027740 & X \\ \hline
\end{tabular}
\end{table}

\newpage

\section{Tables from pairwise comparison for representations}
\label{apdx:statistical_rep}

\begin{table}[htbp]
\centering
\caption{Dunn test results for RNN  - 1s}
\label{dunn_rnn_1s}
\begin{tabular}{|c|c|c|c|c|}
\hline
\multicolumn{5}{|c|}{\textbf{Dunn test for RNN - 1s}} \\
\hline
              & PSDMultitaper & PSDWelch & Spectrogram & Time \\ \hline
PSDMultitaper & X  & 0.108929 & \cellcolor{green}0.004408 & \cellcolor{green}0.000000 \\ \hline
PSDWelch      & 0.108929 & X & 1.000000 & \cellcolor{green}0.004408 \\ \hline
Spectrogram   & \cellcolor{green}0.004408 & 1.000000 & X & 0.108929 \\ \hline
Time          & \cellcolor{green}0.000000 & \cellcolor{green}0.004408 & 0.108929 & X \\ \hline
\end{tabular}
\end{table}

\begin{table}[htbp]
\centering
\caption{Dunn test results for RNN  - 2s}
\label{dunn_rnn_2s}
\begin{tabular}{|c|c|c|c|c|}
\hline
\multicolumn{5}{|c|}{\textbf{Dunn test for RNN - 2s}} \\
\hline
              & PSDMultitaper & PSDWelch & Spectrogram & Time \\ \hline
PSDMultitaper & X & 0.151327 & \cellcolor{green}0.001800 & \cellcolor{green}0.000000 \\ \hline
PSDWelch      & 0.151327 & X & 1.000000 & \cellcolor{green}0.004260 \\ \hline
Spectrogram   & \cellcolor{green}0.001800 & 1.000000 & X & 0.267566 \\ \hline
Time          & \cellcolor{green}0.000000 & \cellcolor{green}0.004260 & 0.267566 & X \\ \hline
\end{tabular}
\end{table}

\begin{table}[htbp]
\centering
\caption{Dunn test results for RNN  - 5s}
\label{dunn_rnn_5s}
\begin{tabular}{|c|c|c|c|c|}
\hline
\multicolumn{5}{|c|}{\textbf{Dunn test for RNN - 5s}} \\
\hline
              & PSDMultitaper & PSDWelch & Spectrogram & Time \\ \hline
PSDMultitaper & X & 0.334420 & \cellcolor{green}0.000722 & \cellcolor{green}0.000000 \\ \hline
PSDWelch      & 0.334420 & X & 0.319994 & \cellcolor{green}0.000844 \\ \hline
Spectrogram   & \cellcolor{green}0.000722 & 0.319994 & X & 0.364899 \\ \hline
Time          & \cellcolor{green}0.000000 & \cellcolor{green}0.000844 & 0.364899 & X \\ \hline
\end{tabular}
\end{table}

\begin{table}[htbp]
\centering
\caption{Dunn test results for RNN  - 10s}
\label{dunn_rnn_10s}
\begin{tabular}{|c|c|c|c|c|}
\hline
\multicolumn{5}{|c|}{\textbf{Dunn test for RNN - 10s}} \\
\hline
              & PSDMultitaper & PSDWelch & Spectrogram & Time \\ \hline
PSDMultitaper & X & 0.397721 & \cellcolor{green}0.000276 & \cellcolor{green}0.000000 \\ \hline
PSDWelch      & 0.397721 & X & 0.151244 & \cellcolor{green}0.002083 \\ \hline
Spectrogram   & \cellcolor{green}0.000276 & 0.151244 & X & 1.000000 \\ \hline
Time          & \cellcolor{green}0.000000 & \cellcolor{green}0.002083 & 1.000000 & X \\ \hline
\end{tabular}
\end{table}

\begin{table}[htbp]
\centering
\caption{Dunn test results for CNN - 1s}
\label{dunn_cnn_1s}
\begin{tabular}{|c|c|c|c|c|}
\hline
\multicolumn{5}{|c|}{\textbf{Dunn test for CNN - 1s}} \\
\hline
 & PSDMultitaper & PSDWelch & Spectrogram & Time \\\hline
PSDMultitaper & X & 0.151203 & \cellcolor{green}0.002497 & \cellcolor{green}0.000000 \\\hline
PSDWelch      & 0.151203 & X & 1.000000 & \cellcolor{green}0.003097 \\\hline
Spectrogram   & \cellcolor{green}0.002497 & 1.000000 & X & 0.175135 \\\hline
Time          & \cellcolor{green}0.000000 & \cellcolor{green}0.003097 & 0.175135 & X \\\hline
\end{tabular}
\end{table}

\begin{table}[htbp]
\centering
\caption{Dunn test results for CNN - 2s}
\label{dunn_cnn_2s}
\begin{tabular}{|c|c|c|c|c|}
\hline
\multicolumn{5}{|c|}{\textbf{Dunn test for CNN - 2s}} \\
\hline
 & PSDMultitaper & PSDWelch & Spectrogram & Time \\\hline
PSDMultitaper & X & \cellcolor{green}0.010210 & 0.055582 & \cellcolor{green}0.000000 \\\hline
PSDWelch      & \cellcolor{green}0.010210 & X & 1.000000 & 0.055582 \\\hline
Spectrogram   & 0.055582 & 1.000000 & X & \cellcolor{green}0.010210 \\\hline
Time          & \cellcolor{green}0.000000 & 0.055582 & \cellcolor{green}0.010210 & X \\\hline
\end{tabular}
\end{table}

\begin{table}[htbp]
\centering
\caption{Dunn test results for CNN - 5s}
\label{dunn_cnn_5s}
\begin{tabular}{|c|c|c|c|c|}
\hline
\multicolumn{5}{|c|}{\textbf{Dunn test for CNN - 5s}} \\
\hline
 & PSDMultitaper & PSDWelch & Spectrogram & Time \\\hline
PSDMultitaper & X & 0.334213 & \cellcolor{green}0.000002 & \cellcolor{green}0.000054 \\\hline
PSDWelch      & 0.334213 & X & \cellcolor{green}0.007842 & 0.069294 \\\hline
Spectrogram   & \cellcolor{green}0.000002 & \cellcolor{green}0.007842 & X & 1.000000 \\\hline
Time          & \cellcolor{green}0.000054 & 0.069294 & 1.000000 & X \\\hline
\end{tabular}
\end{table}

\begin{table}[htbp]
\centering
\caption{Dunn test results for CNN - 10s}
\label{dunn_cnn_10s}
\begin{tabular}{|c|c|c|c|c|}
\hline
\multicolumn{5}{|c|}{\textbf{Dunn test for CNN - 10s}} \\
\hline
 & PSDMultitaper & PSDWelch & Spectrogram & Time \\\hline
PSDMultitaper & X & 0.334351 & \cellcolor{green}0.00057 & \cellcolor{green}0.000000 \\\hline
PSDWelch      & 0.334351 & X & 0.27974 & \cellcolor{green}0.001062 \\\hline
Spectrogram   & \cellcolor{green}0.00057 & 0.27974 & X & 0.47032 \\\hline
Time          & \cellcolor{green}0.000000 & \cellcolor{green}0.001062 & 0.47032 & X \\\hline
\end{tabular}
\end{table}

\begin{table}[htbp]
\centering
\caption{Dunn test results for CRNN - 1s}
\label{dunn_crnn_1s}
\begin{tabular}{|c|c|c|c|c|}
\hline
\multicolumn{5}{|c|}{\textbf{Dunn test for CRNN - 1s}} \\
\hline
 & PSDMultitaper & PSDWelch & Spectrogram & Time \\\hline
PSDMultitaper & X & 0.103371 & \cellcolor{green}0.004403 & \cellcolor{green}0.000000 \\\hline
PSDWelch      & 0.103371 & X & 1.000000 & \cellcolor{green}0.005057 \\\hline
Spectrogram   & \cellcolor{green}0.004403 & 1.000000 & X & 0.114612 \\\hline
Time          & \cellcolor{green}0.000000 & \cellcolor{green}0.005057 & 0.114612 & X \\\hline
\end{tabular}
\end{table}

\begin{table}[htbp]
\centering
\caption{Dunn test results for CRNN - 2s}
\label{dunn_crnn_2s}
\begin{tabular}{|c|c|c|c|c|}
\hline
\multicolumn{5}{|c|}{\textbf{Dunn test for CRNN - 2s}} \\
\hline
 & PSDMultitaper & PSDWelch & Spectrogram & Time \\\hline
PSDMultitaper & X & \cellcolor{green}0.001238 & 0.233004 & \cellcolor{green}0.000000 \\\hline
PSDWelch      & \cellcolor{green}0.001238 & X & 0.599692 & 0.279925 \\\hline
Spectrogram   & 0.233004 & 0.599692 & X & \cellcolor{green}0.001671 \\\hline
Time          & \cellcolor{green}0.000000 & 0.279925 & \cellcolor{green}0.001671 & X \\\hline
\end{tabular}
\end{table}

\begin{table}[htbp]
\centering
\caption{Dunn test results for CRNN - 5s}
\label{dunn_crnn_5s}
\begin{tabular}{|c|c|c|c|c|}
\hline
\multicolumn{5}{|c|}{\textbf{Dunn test for CRNN - 5s}} \\
\hline
 & PSDMultitaper & PSDWelch & Spectrogram & Time \\\hline
PSDMultitaper & X & 0.334006 & \cellcolor{green}0.000000 & \cellcolor{green}0.000275 \\\hline
PSDWelch      & 0.334006 & X & \cellcolor{green}0.002074 & 0.183524 \\\hline
Spectrogram   & \cellcolor{green}0.000000 & \cellcolor{green}0.002074 & X & 0.940504 \\\hline
Time          & \cellcolor{green}0.000275 & 0.183524 & 0.940504 & X \\\hline
\end{tabular}
\end{table}

\begin{table}[htbp]
\centering
\caption{Dunn test results for CRNN - 10s}
\label{dunn_crnn_10s}
\begin{tabular}{|c|c|c|c|c|}
\hline
\multicolumn{5}{|c|}{\textbf{Dunn test for CRNN - 10s}} \\
\hline
 & PSDMultitaper & PSDWelch & Spectrogram & Time \\\hline
PSDMultitaper & X & 0.432693 & \cellcolor{green}0.000000 & \cellcolor{green}0.000527 \\\hline
PSDWelch      & 0.432693 & X & \cellcolor{green}0.001146 & 0.202207 \\\hline
Spectrogram   & \cellcolor{green}0.000000 & \cellcolor{green}0.001146 & X & 0.648234 \\\hline
Time          & \cellcolor{green}0.000527 & 0.202207 & 0.648234 & X \\\hline
\end{tabular}
\end{table}

\newpage

\section{Tables from pairwise comparison for neural network architecture}
\label{apdx:statistical_arc}

\begin{table}[htbp]
\centering
\caption{Friedman test results for Time - 1s}
\label{friedman_time_1s}
\begin{tabular}{|c|c|c|c|}
\hline
\multicolumn{4}{|c|}{\textbf{Friedman test for Time - 1s}} \\
\hline
 & CNN & CRNN & RNN \\\hline
CNN  & X & 0.9919 & \cellcolor{green}0.0001 \\\hline
CRNN & 0.9919 & X & \cellcolor{green}0.0001 \\\hline
RNN  & \cellcolor{green}0.0001 & \cellcolor{green}0.0001 & X \\\hline
\end{tabular}
\end{table}

\begin{table}[htbp]
\centering
\caption{Friedman test results for Time - 2s}
\label{friedman_time_2s}
\begin{tabular}{|c|c|c|c|}
\hline
\multicolumn{4}{|c|}{\textbf{Friedman test for Time - 2s}} \\
\hline
 & CNN & CRNN & RNN \\\hline
CNN  & X & \cellcolor{green}0.001 & \cellcolor{green}0.0 \\\hline
CRNN & \cellcolor{green}0.001 & X & \cellcolor{green}0.0 \\\hline
RNN  & \cellcolor{green}0.0 & \cellcolor{green}0.0 & X \\\hline
\end{tabular}
\end{table}

\begin{table}[htbp]
\centering
\caption{Friedman test results for Time - 5s}
\label{friedman_time_5s}
\begin{tabular}{|c|c|c|c|}
\hline
\multicolumn{4}{|c|}{\textbf{Friedman test for Time - 5s}} \\
\hline
 & CNN & CRNN & RNN \\\hline
CNN  & X & \cellcolor{green}0.0082 & \cellcolor{green}0.0 \\\hline
CRNN & \cellcolor{green}0.0082 & X & \cellcolor{green}0.0 \\\hline
RNN  & \cellcolor{green}0.0 & \cellcolor{green}0.0 & X \\\hline
\end{tabular}
\end{table}

\begin{table}[htbp]
\centering
\caption{Friedman test results for Time - 10s}
\label{friedman_time_10s}
\begin{tabular}{|c|c|c|c|}
\hline
\multicolumn{4}{|c|}{\textbf{Friedman test for Time - 10s}} \\
\hline
 & CNN & CRNN & RNN \\\hline
CNN  & X & \cellcolor{green}0.0 & \cellcolor{green}0.0 \\\hline
CRNN & \cellcolor{green}0.0 & X & \cellcolor{green}0.0 \\\hline
RNN  & \cellcolor{green}0.0 & \cellcolor{green}0.0 & X \\\hline
\end{tabular}
\end{table}

\begin{table}[htbp]
\centering
\caption{Friedman test results for PSDWelch - 2s}
\label{friedman_psdwelch_2s}
\begin{tabular}{|c|c|c|c|}
\hline
\multicolumn{4}{|c|}{\textbf{Friedman test for PSDWelch - 2s}} \\
\hline
 & CNN & CRNN & RNN \\\hline
CNN  & X & \cellcolor{green}0.0 & \cellcolor{green}0.0 \\\hline
CRNN & \cellcolor{green}0.0 & X & \cellcolor{green}0.0 \\\hline
RNN  & \cellcolor{green}0.0 & \cellcolor{green}0.0 & X \\\hline
\end{tabular}
\end{table}

\begin{table}[htbp]
\centering
\caption{Friedman test results for PSDWelch - 5s}
\label{friedman_psdwelch_5s}
\begin{tabular}{|c|c|c|c|}
\hline
\multicolumn{4}{|c|}{\textbf{Friedman test for PSDWelch - 5s}} \\
\hline
 & CNN & CRNN & RNN \\\hline
CNN  & X & \cellcolor{green}0.0008 & \cellcolor{green}0.0 \\\hline
CRNN & \cellcolor{green}0.0008 & X & \cellcolor{green}0.0418 \\\hline
RNN  & \cellcolor{green}0.0 & \cellcolor{green}0.0418 & X \\\hline
\end{tabular}
\end{table}

\begin{table}[htbp]
\centering
\caption{Friedman test results for PSDWelch - 10s}
\label{friedman_psdwelch_10s}
\begin{tabular}{|c|c|c|c|}
\hline
\multicolumn{4}{|c|}{\textbf{Friedman test for PSDWelch - 10s}} \\
\hline
 & CNN & CRNN & RNN \\\hline
CNN  & X & \cellcolor{green}0.0405 & 0.7126 \\\hline
CRNN & \cellcolor{green}0.0405 & X & \cellcolor{green}0.0063 \\\hline
RNN  & 0.7126 & \cellcolor{green}0.0063 & X \\\hline
\end{tabular}
\end{table}

\begin{table}[htbp]
\centering
\caption{Friedman test results for PSDMultitaper - 1s}
\label{friedman_psdmultitaper_1s}
\begin{tabular}{|c|c|c|c|}
\hline
\multicolumn{4}{|c|}{\textbf{Friedman test for PSDMultitaper - 1s}} \\
\hline
 & CNN & CRNN & RNN \\\hline
CNN  & X & 0.2237 & \cellcolor{green}0.0022 \\\hline
CRNN & 0.2237 & X & 0.114 \\\hline
RNN  & \cellcolor{green}0.0022 & 0.114 & X \\\hline
\end{tabular}
\end{table}

\begin{table}[htbp]
\centering
\caption{Friedman test results for PSDMultitaper - 2s}
\label{friedman_psdmultitaper_2s}
\begin{tabular}{|c|c|c|c|}
\hline
\multicolumn{4}{|c|}{\textbf{Friedman test for PSDMultitaper - 2s}} \\
\hline
 & CNN & CRNN & RNN \\\hline
CNN  & X & \cellcolor{green}0.0 & \cellcolor{green}0.0 \\\hline
CRNN & \cellcolor{green}0.0 & X & \cellcolor{green}0.0036 \\\hline
RNN  & \cellcolor{green}0.0 & \cellcolor{green}0.0036 & X \\\hline
\end{tabular}
\end{table}

\begin{table}[htbp]
\centering
\caption{Friedman test results for PSDMultitaper - 5s}
\label{friedman_psdmultitaper_5s}
\begin{tabular}{|c|c|c|c|}
\hline
\multicolumn{4}{|c|}{\textbf{Friedman test for PSDMultitaper - 5s}} \\
\hline
 & CNN & CRNN & RNN \\\hline
CNN  & X & \cellcolor{green}0.0 & \cellcolor{green}0.0 \\\hline
CRNN & \cellcolor{green}0.0 & X & 0.6516 \\\hline
RNN  & \cellcolor{green}0.0 & 0.6516 & X \\\hline
\end{tabular}
\end{table}

\begin{table}[htbp]
\centering
\caption{Friedman test results for PSDMultitaper - 10s}
\label{friedman_psdmultitaper_10s}
\begin{tabular}{|c|c|c|c|}
\hline
\multicolumn{4}{|c|}{\textbf{Friedman test for PSDMultitaper - 10s}} \\
\hline
 & CNN & CRNN & RNN \\\hline
CNN  & X & \cellcolor{green}0.0 & \cellcolor{green}0.0001 \\\hline
CRNN & \cellcolor{green}0.0 & X & 0.2961 \\\hline
RNN  & \cellcolor{green}0.0001 & 0.2961 & X \\\hline
\end{tabular}
\end{table}

\begin{table}[htbp]
\centering
\caption{Friedman test results for Spectrogram - 2s}
\label{friedman_spectrogram_2s}
\begin{tabular}{|c|c|c|c|}
\hline
\multicolumn{4}{|c|}{\textbf{Friedman test for Spectrogram - 2s}} \\
\hline
 & CNN & CRNN & RNN \\\hline
CNN  & X & 0.1914 & \cellcolor{green}0.0 \\\hline
CRNN & 0.1914 & X & \cellcolor{green}0.0 \\\hline
RNN  & \cellcolor{green}0.0 & \cellcolor{green}0.0 & X \\\hline
\end{tabular}
\end{table}

\begin{table}[htbp]
\centering
\caption{Friedman test results for Spectrogram - 5s}
\label{friedman_spectrogram_5s}
\begin{tabular}{|c|c|c|c|}
\hline
\multicolumn{4}{|c|}{\textbf{Friedman test for Spectrogram - 5s}} \\
\hline
 & CNN & CRNN & RNN \\\hline
CNN  & X & 0.8278 & \cellcolor{green}0.0 \\\hline
CRNN & 0.8278 & X & \cellcolor{green}0.0 \\\hline
RNN  & \cellcolor{green}0.0 & \cellcolor{green}0.0 & X \\\hline
\end{tabular}
\end{table}

\begin{table}[htbp]
\centering
\caption{Friedman test results for Spectrogram - 10s}
\label{friedman_spectrogram_10s}
\begin{tabular}{|c|c|c|c|}
\hline
\multicolumn{4}{|c|}{\textbf{Friedman test for Spectrogram - 10s}} \\
\hline
 & CNN & CRNN & RNN \\\hline
CNN  & X & 0.9033 & \cellcolor{green}0.0 \\\hline
CRNN & 0.9033 & X & \cellcolor{green}0.0 \\\hline
RNN  & \cellcolor{green}0.0 & \cellcolor{green}0.0 & X \\\hline
\end{tabular}
\end{table}

\newpage

 \bibliographystyle{elsarticle-num} 
 \bibliography{cas-refs}





\end{document}